\documentclass{article}
\usepackage{arxiv}

\usepackage[utf8]{inputenc} 
\usepackage[T1]{fontenc}    
\usepackage{palatino}
\usepackage{hyperref}       
\usepackage{url}            
\usepackage{booktabs}       
\usepackage{amsfonts}       
\usepackage{nicefrac}       
\usepackage{microtype}      
\usepackage{xcolor}         

\definecolor{darkblue}{rgb}{0, 0, 0.5}
\hypersetup{
    colorlinks=true,
    citecolor=darkblue,
    linkcolor=darkblue,
    urlcolor=darkblue,
}

\usepackage{amsmath,amsfonts,bm}

\def\eqref#1{equation~\ref{#1}}

\def\1{\bm{1}}

\DeclareMathAlphabet{\mathsfit}{\encodingdefault}{\sfdefault}{m}{sl}
\SetMathAlphabet{\mathsfit}{bold}{\encodingdefault}{\sfdefault}{bx}{n}

\usepackage[verbose=true,letterpaper]{geometry}
\usepackage{natbib}
\usepackage[hyphenbreaks]{breakurl} 
\usepackage{float}
\usepackage{xspace}
\usepackage{enumerate}
\usepackage{graphicx}
\usepackage{pgf}
\usepackage{paralist}
\usepackage{wrapfig}
\usepackage{nicefrac}       

\setcitestyle{authoryear,open={(},close={)}}

\usepackage{fullpage}

\usepackage{graphicx}
\usepackage{bbm}
\usepackage{enumitem}
\usepackage{multirow}
\usepackage{setspace}
\usepackage{cleveref}

\usepackage{amsmath}
\usepackage{amssymb}
\usepackage{mathtools}
\usepackage{amsthm}

\theoremstyle{plain}

\theoremstyle{remark}

\usepackage{listings}
\usepackage{siunitx}
\usepackage{bookmark}
\usepackage{textcomp}
\usepackage{nth}
\usepackage{indentfirst}
\usepackage{siunitx}
\usepackage{changepage}
\usepackage{bm}
\usepackage{multirow}
\usepackage{multicol}
\usepackage{colortbl}

\usepackage{subfig}
\usepackage{makecell}

\usepackage{titling}
\usepackage{setspace}

\usepackage[most,skins,theorems,breakable]{tcolorbox}
\tcbset{
  mybox/.style={
    width=\linewidth,
    top=14pt,
    bottom=7pt,
    colback=cyan!5!white,
    colframe=black,
    colbacktitle=black,
    enhanced,
    center,
    breakable,
    attach boxed title to top left={yshift=-0.13in,xshift=0.15in},
    boxed title style={boxrule=0pt,colframe=white,},
  }
}
\newtcolorbox{mybox}[2][]{mybox,title=#2,#1}

\usepackage{algorithm}
\usepackage{algorithmicx}
\usepackage{algpseudocode}

\newcommand{\ourmethod}{\textcolor{softblue}{\texttt{Rejuvenation}}\xspace}
\newcommand{\STR}{SFT-then-RL\xspace}
\newcommand{\base}{\textcolor{softgray}{\texttt{Base}}\xspace}
\newcommand{\under}{\textcolor{softyellow}{\texttt{UnderSFT}}\xspace}
\renewcommand{\mod}{\textcolor{softgreen}{\texttt{ModSFT}}\xspace}
\renewcommand{\over}{\textcolor{softred}{\texttt{OverSFT}}\xspace}
\newcommand{\dft}{\texttt{DFT}\xspace}

\newcommand{\TB}{$\tau$-bench\xspace}

\newcommand{\fourB}{EvoLM-4B\xspace}
\newcommand{\eightB}{Qwen3-8B\xspace}

\colorlet{softred}{red!80!black}
\colorlet{softgreen}{green!55!black}
\colorlet{softblue}{blue!65!black}
\colorlet{softyellow}{yellow!70!black}
\colorlet{softgray}{black!50}
\definecolor{softblue}{RGB}{38, 139, 210}

\usepackage{array}
\newcolumntype{L}[1]{>{\raggedright\arraybackslash}p{#1}}
\newcolumntype{C}[1]{>{\centering\arraybackslash}p{#1}}

\newcommand\blfootnote[1]{%
  \begingroup
  \renewcommand\thefootnote{}\footnote{#1}%
  \addtocounter{footnote}{-1}%
  \endgroup
}

\title{\textbf{When RL Fails after SFT: Rejuvenating Model Plasticity\\ for Robust SFT-to-RL Handoff}}
\author{
{{\textbf{Runze Liu$^{1*}$ ~ Jiashun Liu$^{1*}$ ~ Xu Wan$^{2}$ ~ Yuqian Fu$^{3}$ ~ Ling Pan$^{1}$}}}\\
{\normalsize{$^{1}$Hong Kong University of Science and Technology $^{2}$Zhejiang University}}\\
{\normalsize{$^{3}$State Key Laboratory of Multimodal Artificial Intelligence Systems, CASIA}}\\
}
\date{}

\begin{document}

\maketitle

\vspace{-3.0em}
\blfootnote{$^*$ Equal contribution}

\begin{abstract}
Supervised Fine-Tuning (SFT) followed by Reinforcement Learning (RL) has become a standard pipeline for Large Language Model (LLM) post-training. SFT is expected to provide a useful behavioral prior for RL to further enhance model capabilities. However, checkpoints with excessive SFT often show limited improvement during RL. We attribute this failure to the loss of model plasticity: the reduced ability of an SFT-initialized policy to be effectively reshaped by subsequent RL. To better understand this phenomenon, we conduct detailed analysis from multiple perspectives, including parameter changes, output spaces, and RL optimization dynamics. Our results show that models from excessive SFT tend to produce over-confident token distributions and exhibit sharp parameter landscapes, which make them harder to optimize in the RL stage. To enable a more robust SFT-to-RL handoff, we propose \texttt{Rejuvenation}, a simple yet effective method that restores plasticity while preserving useful SFT-acquired priors. Rejuvenation leverages base-anchored model fusion to reduce excessive SFT-induced drift with targeted neuron reset to mitigate model rigidity. Experimental results on both math reasoning tasks and agentic tasks demonstrate that our approach consistently improves RL performance on over-trained SFT models, while also enhancing generalization to out-of-distribution tasks.
\end{abstract}

\section{Introduction}
\label{sec:introduction}

Post-training has emerged as a critical phase for unlocking the reasoning and agentic capabilities of large language models (LLMs)~\citep{o1, DeepSeek-R1}. Many practical systems adopt the SFT-then-RL~\citep{DeepSeek-R1} pipeline, where supervised fine-tuning (SFT)~\citep{ouyang2022training, bai2022training} first teaches models to follow instructions, produce the desired format, and acquire cold-start knowledge, after which reinforcement learning (RL)~\citep{GRPO, zhang2025survey} further optimizes the policy according to task rewards~\citep{Limit-of-RLVR, ProRL, ASPO, Archer, AttnRL, MARTI}. In this pipeline, SFT is not merely for imitating high-quality solutions, but also determines the initialization that RL inherits. This pipeline therefore relies on an implicit assumption: after acquiring useful behaviors through imitation, the SFT checkpoint should still serve as a suitable starting point for reward-driven optimization.

\begin{figure*}[!t]
\centering
\includegraphics[width=1.0\textwidth]{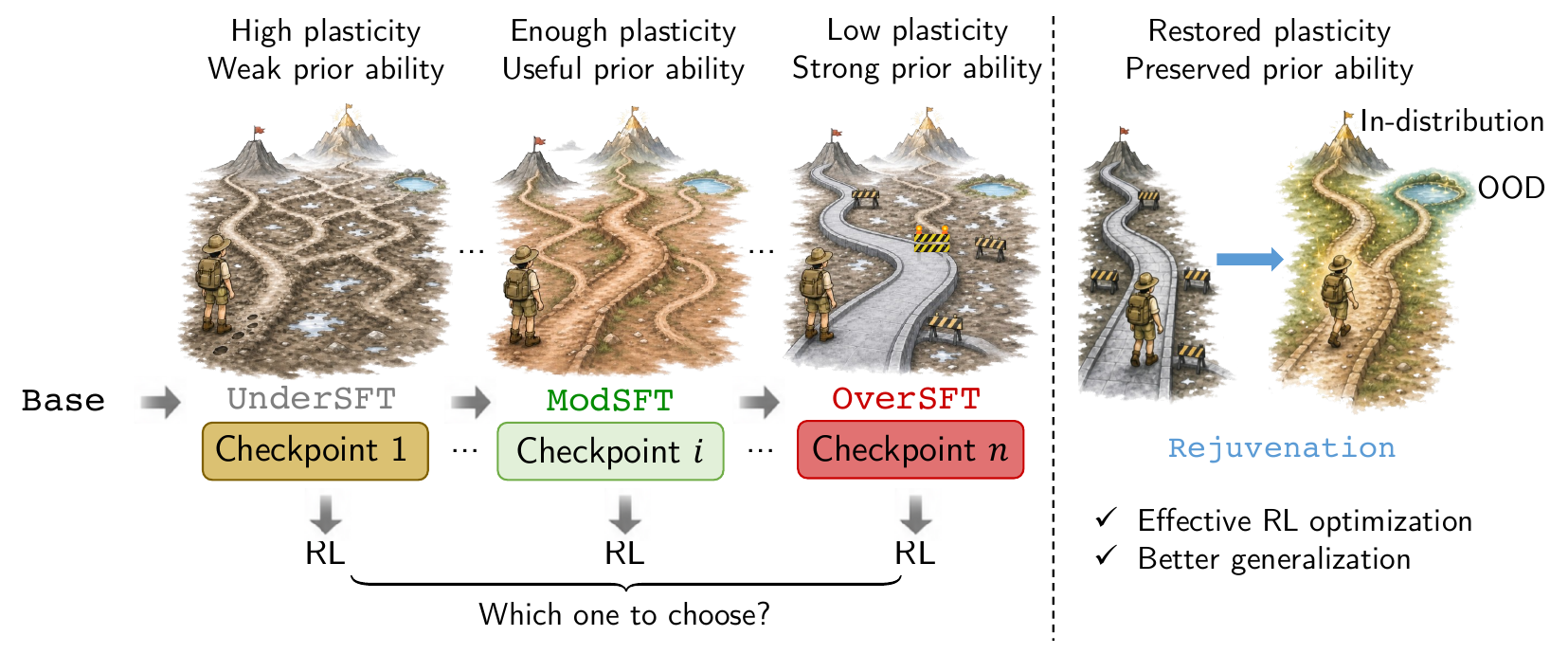}
\caption{Overview of \ourmethod.}
\label{fig:overview}
\end{figure*}

However, the amount of SFT needed for strong supervised behavior may not coincide with the amount of SFT that yields the best final model after RL. Intuitively, if SFT is stopped too early, the model may be under-prepared for RL, lacking the instruction-following patterns or task-specific behaviors needed for efficient optimization. Meanwhile, if SFT is continued for too long, the model may become overly specialized to the supervised data, and the resulting checkpoint may leave limited room for RL to further improve the policy and generalize~\citep{chu2025sft}. Thus, despite its widespread success, the handoff from SFT to RL remains a fragile and computationally expensive. A critical yet often overlooked challenge is to determine which SFT checkpoint should be used as the RL initial policy.

Standard SFT metrics (e.g., training loss, validation accuracy) measure imitation quality, not the checkpoint's capacity for reward-driven improvement. A natural remedy is early stopping~\citep{AESL}, but it still assumes that the appropriate stopping point can be reliably identified, and therefore does not resolve the core problem. Recent work improves the SFT stage through better data~\citep{huang2026fine, GenPRM}, objectives~\citep{SRFT, PSFT}, or training recipes~\citep{PRMLessons}, but these methods primarily change the SFT trajectory rather than provide a reliable criterion for RL readiness. As a result, practitioners still often have to launch RL from multiple SFT checkpoints to determine the appropriate handoff point. This search is expensive, and its outcome can be sensitive to RL hyperparameters and optimization noise.

In this paper, we aim to analyze and address this critical dilemma in the SFT-to-RL handoff, where insufficient SFT may prematurely stop imitation learning that prevents the model from acquiring useful skills, while over-trained SFT becomes highly resistant to further improvement via RL. To better understand this issue, we conduct a detailed analysis of how extended SFT changes both model behavior and subsequent optimization dynamics under RL. We find that \over models tend to produce sharper, more over-confident output distributions and less smooth parameter space, showing large gradient norms but limited performance gains and smaller parameter update magnitude compared to \mod models. These findings suggest that the excessive SFT does not merely overfit the supervised data, but also make the policy resistant and becomes less effectively adaptable in the subsequent RL stage. We identify this failure mode as a loss of \textit{model plasticity}\footnote{We use model plasticity to refer to the ability of SFT models to undergo reward-driven improvement in subsequent RL. A plastic model should remain responsive to RL update and such updates can effectively translate into task performance gains.}~\citep{dohare2024loss, han2026weight}: the model becomes difficult to reshape through RL.

Based on this observation, we propose \ourmethod, a simple yet effective post-hoc mechanism that enables robust SFT-to-RL handoff for recovering model plasticity, which avoids complex SFT loss modifications and costly checkpoint searches. 
Our key insight is that SFT should provide a useful behavioral prior, but not at the cost of losing plasticity.
Our approach is a dual-level strategy. First, at the global level, we utilize base-anchored model fusion to reduce excessive SFT-induced drift while retaining useful behavior.
At the local level, we introduce a targeted neuron reset mechanism based on logit attribution to selectively restore diversity in over-confident LLM most responsible for collapsed predictions.
As shown in Figure~\ref{fig:overview}, our method effectively mitigates the resulting rigidity induced by excessive SFT, while preserving the effective behavioral prior acquired from sufficient SFT.
We evaluate \ourmethod on both mathematical reasoning tasks and agentic tasks. Experiments show that our method not only consistently recovers RL improvement from previously \over models, but also achieves superior performance compared to \mod models on out-of-distribution (OOD) tasks with even better generalization ability.

The main contributions of this work can be summarized as follows:

\begin{enumerate}[leftmargin=*, topsep=0pt, parsep=0pt, partopsep=0pt]
    \item We identify a failure mode in the SFT-then-RL pipeline: the SFT-to-RL handoff dilemma, where fully-trained SFT models lose plasticity and limit RL improvement.
    \item We provide detailed analysis from multiple perspectives, revealing that over-SFT leads to reduced effective gradients during RL, which leads to entropy collapse and fundamentally hurts RL optimization dynamics.
    \item We propose a simple, cheap, and effective rejuvenation method to recover model plasticity post-hoc using model fusion and neuron reset, making it robust to different SFT-to-RL handoffs.
    \item We demonstrate the effectiveness of our method on both mathematical and agentic tasks, showing that it consistently recovers RL improvement from \over checkpoints and improves OOD generalization over \mod baselines.
\end{enumerate}

\section{Related Work}
\label{sec:related_work}

\paragraph{SFT and RL in LLM Post-Training.}
Recent work has shown that RL emerges as an effective method for LLM post-training~\citep{o1, DeepSeek-R1, GRPO, DAPO}. Many methods aim to better integrate SFT and RL, such as improving the use of off-policy data~\citep{LUFFY, SASR, SuperRL, ReLIFT, Prefix-RFT}, designing unified training objectives~\citep{UFT, SRFT, Chord, HPT, GFT}, or using importance weighted objectives to better align SFT with RL optimization~\citep{PSFT, iw-SFT, PEAR}. At the same time, recent evidence suggests that comparisons between mixed-policy methods and the standard \STR pipeline can be sensitive to SFT implementation details, and that carefully controlled \STR remains a strong baseline~\citep{SFT-then-RL}. Several recent studies further analyze why SFT and RL lead to different generalization behavior: SFT tends to memorize supervised data while RL can improve out-of-distribution generalization~\citep{chu2025sft}, RL may partially heal OOD forgetting introduced by SFT but only within a suitable checkpoint range~\citep{jin2025rl1, jin2025rl2}, and high SFT scores are not necessarily reliable predictors of post-RL performance~\citep{kang2026quagmires, PEAR}. These works expose the fragility of the SFT-to-RL handoff, but they mainly diagnose checkpoint selection or redesign the SFT objective. In contrast, we ask whether the plasticity of an already over-trained SFT model can be restored post-hoc before RL starts.

\paragraph{Overfitting and Regularization in SFT.}

Recently, many works have explored how to prevent overfitting or excessive policy drift during SFT. Methods such as GEM~\citep{GEM}, PSFT~\citep{PSFT}, ASFT~\citep{ASFT}, and CurioSFT~\citep{CurioSFT}, introduce an auxiliary regularization loss to maintain model diversity. DFT~\citep{DFT}, AESL~\citep{AESL}, and ProFit~\citep{ProFit} incorporate probability-based weighting in the cross-entropy loss. However, they mainly focus on preventing overfitting during SFT or designing a better SFT objective from the beginning. 
Our setting is different that we assume an over-trained SFT model has already been obtained, and ask whether its plasticity can be restored for subsequent RL.

\section{Diagnosing and Restoring Plasticity in Over-Trained Models}
\label{sec:analysis}

In this section, we analyze why over-trained SFT models become difficult to improve with RL and then introduce two post-hoc recovery operations. 
We first investigate what excessive SFT changes in both parameter and output spaces before any RL update is applied, to understand the handoff-specific question: \emph{does continued SFT move the checkpoint into a state that is less amemnable to subsequent RL optimization?}
We then connect these changes to poor RL trainability, where large gradient norms do not translate into effective parameter movement or meaningful performance gains (Section~\ref{sec:analysis_rl}). Motivated by these observations, we
recover plasticity at two levels: base-anchored model fusion globally pulls the model toward a smoother region (Section~\ref{sec:analysis_fusion}), while attribution-guided neuron reset locally restores the high-contribution directions
responsible for abnormal logits (Section~\ref{sec:analysis_reset}).

\subsection{How Does Over-Training Change the SFT Model?}
\label{sec:analysis_sft}

\subsubsection{Parameter Space}

We train EvoLM-4B~\citep{EvoLM} on the math SFT data and save checkpoints along the SFT process. We denote the moderately trained checkpoint (epoch=2) as \mod and the over-trained checkpoint (epoch=32) as \over. More training details are provided in Section~\ref{sec:exp_setup} and Appendix~\ref{app:exp_setup}.

\begin{wrapfigure}[14]{r}{0.4\textwidth}
\centering
\small
\vspace{-1.5em}
\includegraphics[width=0.4\textwidth]{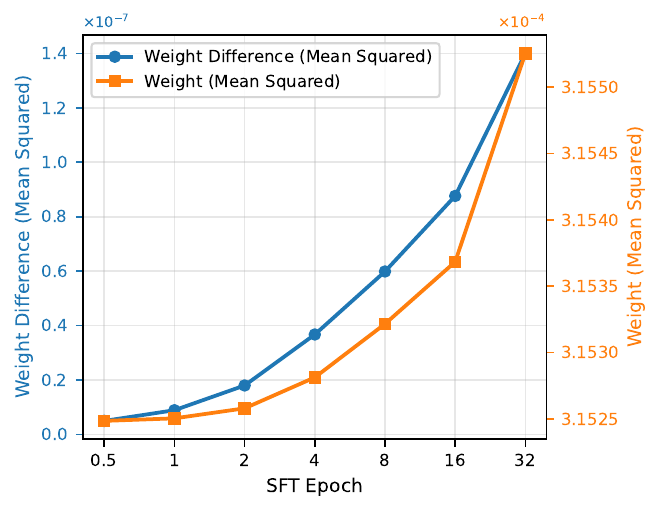}
\caption{Parameter changes and statistics of different SFT checkpoints.}
\label{fig:param_change_sft}
\end{wrapfigure}

\begin{figure*}[!h]
\centering
\vspace{-1.0em}
\includegraphics[width=0.72\textwidth]{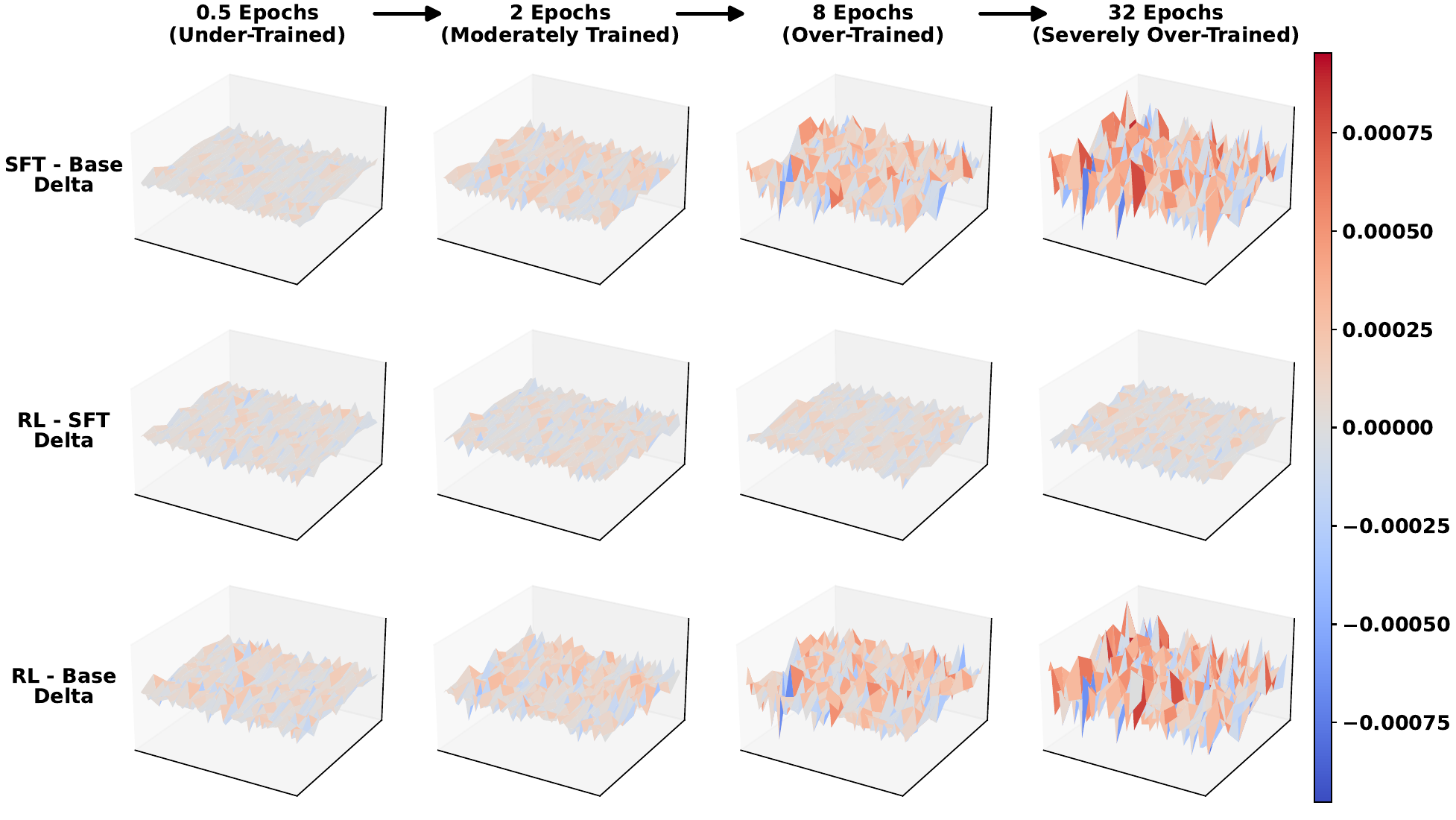}
\caption{Parameter changes of \texttt{layers.0.self\_attn.v\_proj.weight} induced by SFT and RL.}
\vspace{-1.0em}
\label{fig:param_change_l0_vproj}
\end{figure*}

\paragraph{Excessive SFT induces large parameter shifts and weight magnitude.}
We first examine how SFT changes the model parameters. For each checkpoint, we visualize the element-wise parameter difference with respect to the base model. As shown in Figure~\ref{fig:param_change_sft} and the first row of Figure~\ref{fig:param_change_l0_vproj}, we observe that \mod only introduces moderate and relatively smooth parameter changes, while \over leads to extremely large parameter shifts and there are sharp spikes in the shifts, resulting in larger weight magnitude. These spikes indicate that over-training does not simply continue improving the same solution found by moderate SFT. Instead, it drives a small subset of parameters significantly far away from the base model. Additionally, these observations are consistent across all modules and layers. Please find Appendix~\ref{app:exp_results} for more visualizations.

\paragraph{Subsequent RL induces much smaller movement than the preceding SFT-induced drift.}
We further examine how RL changes the model parameters by plotting the difference between: (1) RL vs SFT and (2) RL vs Base. Figure~\ref{fig:param_change_l0_vproj} shows that the subsequent RL stage only introduces much smaller changes compared with the SFT-induced shift. This suggests that once SFT has caused relatively high parameter distortion, RL is unlikely to reverse it through standard policy optimization.

\subsubsection{Output Space}

\paragraph{Excessive SFT saturates the policy before RL.}
We compare the output diversity of different SFT checkpoints in Table~\ref{tab:analysis_output_diversity}. We can observe that \over achieves near-zero training loss but exhibits substantially lower token entropy than \mod (0.024 vs. 0.184), suggesting that excessive SFT drives the policy toward an over-confident regime. 
Although \over attains a higher Pass@1 score than \mod, its Pass@K-Pass@1 gap is the smallest, indicating reduced output diversity despite stronger greedy performance.
Consistently, Figure~\ref{fig:logit_dist} shows logit distributions where \over concentrates probability mass almost entirely on a single token~\citep{GEM}, whereas \mod still assigns non-negligible probability to alternative tokens.

\begin{table*}[!ht]
\centering
\caption{Diversity-related metrics of different SFT checkpoints.}
\resizebox{0.8\textwidth}{!}{
\begin{tabular}{l*{4}{c}}
\toprule
{\textbf{Checkpoint}} & {\textbf{Training Loss}} & {\textbf{Entropy}} & {\textbf{MATH-500 Pass@1}} & {\textbf{Pass@K - Pass@1 Gap}} \\
\midrule
\under & 0.178 & 0.281 & 9.7  & 13.7 \\
\mod   & 0.111 & 0.184 & 15.2 & \textbf{15.3} \\
\over  & 0.002 & 0.024 & \textbf{15.8} & 11.2 \\
\bottomrule
\end{tabular}%
}
\label{tab:analysis_output_diversity}%
\end{table*}%

\begin{mybox}{Takeaway}
Over-trained SFT models differ from moderately trained models in both parameter and output spaces: they contain large parameter shifts and produce over-confident token distributions. These two signatures indicate reduced plasticity before RL starts.
\end{mybox}

\subsection{Does RL Fail after Over-Trained SFT?}
\label{sec:analysis_rl}

\subsubsection{Training Dynamics}

\begin{wrapfigure}[10]{r}{0.36\textwidth}
\centering
\small
\vspace{-3em}
\includegraphics[width=0.36\textwidth]{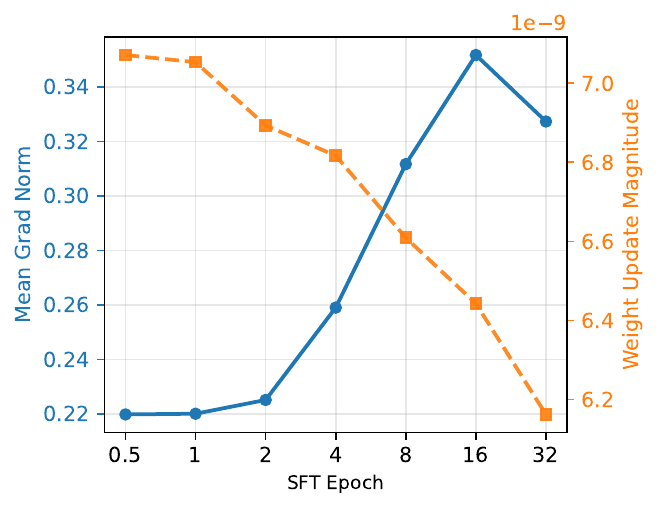}
\caption{Average gradient norm and weight update magnitude of RL with different SFT checkpoints.}
\label{fig:rl_grad_update_mag}
\end{wrapfigure}

\paragraph{Over-trained models exhibit larger RL gradient norms.}
We next study whether these SFT-induced changes affect RL optimization. Figure~\ref{fig:rl_grad_update_mag} shows an apparent paradox: \over has a much larger gradient norm than \mod throughout RL, yet its RL-induced parameter shift is smaller (Figure~\ref{fig:param_change_l0_vproj}, RL-vs-SFT row) and its reward improvement is smaller as well.

\subsubsection{RL Performance Gain}

\begin{figure*}[!t]
\centering
\includegraphics[width=1\textwidth]{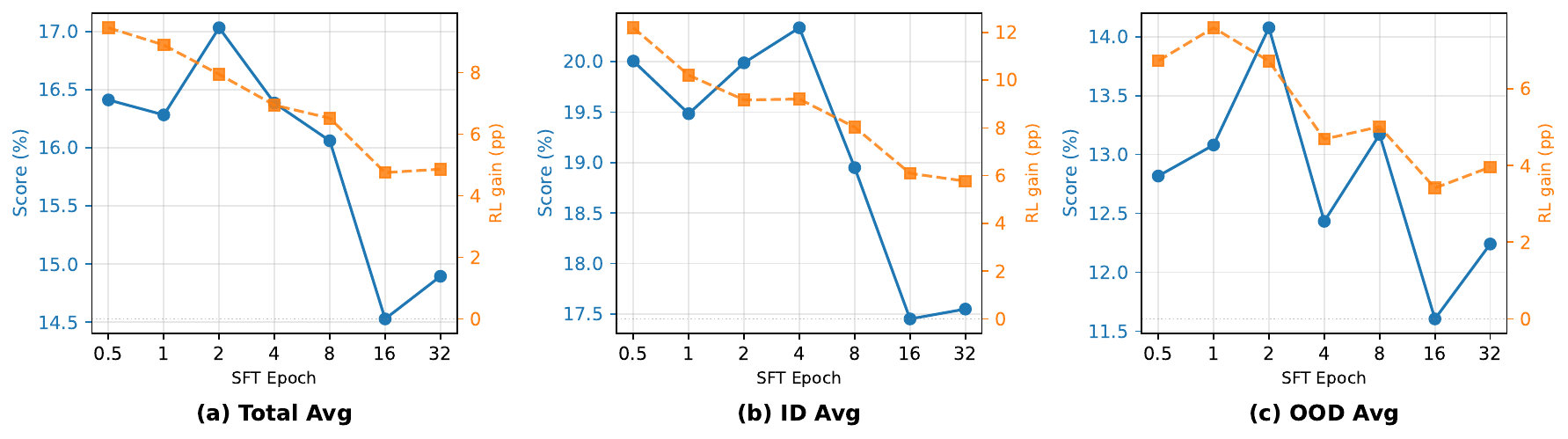}
\caption{Evaluation results of RL-trained models from different SFT checkpoints. (a) Average score of ID and OOD tasks. (b) Average score on ID tasks. (c) Average score on OOD tasks.}
\label{fig:RL_gain}
\end{figure*}

\paragraph{Over-training reduces RL performance gain.}
Figure~\ref{fig:RL_gain} shows that after excessive SFT training, the RL gain of \over models drop significantly compared with \mod models, demonstrating that over-training harms the RL performance gain.

\begin{mybox}{Takeaways}
\begin{itemize}[leftmargin=6pt]
    \item Over-trained models are hard to optimize during RL, exhibiting higher gradient norm but inducing less weight update magnitude.
    \item Over-training leads to smaller performance gains after RL on both ID and OOD tasks.
\end{itemize}
\end{mybox}

\subsection{Global Recovery via Base-Anchored Fusion}
\label{sec:analysis_fusion}

Section~\ref{sec:analysis_sft} shows that over-training pushes a small set of parameters far from the base model and places the model in a sharp region. A natural fix is to pull the over-trained model back toward the base so that the shift shrinks and the surrounding landscape becomes smoother, while still keeping the useful behavior that SFT has acquired~\citep{model-soups, ilharco2023editing}.

\paragraph{Base-anchored linear fusion.}
We therefore perform a simple element-wise linear interpolation between the base model and the \over model:
\begin{equation}
    \theta_{\mathrm{fuse}} = \alpha \, \theta_{\over} + (1 - \alpha) \, \theta_{\base}, \qquad \alpha \in [0, 1].
\label{eq:model_fusion}
\end{equation}
The interpolation is applied to all model parameters, including attention and MLP weights as well as RMSNorm weights, the final norm, and \texttt{lm\_head}. The fusion weight $\alpha$ controls how much of \over is retained: $\alpha \to 1$ reduces to the original \over model, and $\alpha \to 0$ falls back to the base model.

\paragraph{Why fusion helps.}
This operation scales the SFT-induced weight delta $\theta_{\over}-\theta_{\base}$ by $\alpha$. From the task-vector perspective of model editing and composition~\citep{ilharco2023editing}, the SFT update $\theta_{\over}-\theta_{\base}$ can be viewed as a task vector that moves the base model toward the SFT solution. Base-anchored fusion preserves the direction of this task vector while reducing its magnitude, thereby retaining much of the task-specific behavior learned during SFT but avoiding the full parameter displacement caused by over-SFT.
Additionally, fusion shrinks the element-wise deviation from the base to $\alpha \cdot (\theta_{\over} - \theta_{\base})$, which reduces the large shifts and sharp spikes in Figure~\ref{fig:param_change_l0_vproj}. The fused model therefore sits in a smoother region that is easier for RL to optimize. In practice, a moderate $\alpha$ (e.g., $\alpha=0.5$) recovers most of the training-dynamics signals and token entropy while preserving much of the ID performance gained from SFT.

\subsection{Local Recovery via Attribution-Guided Neuron Reset}
\label{sec:analysis_reset}

Global fusion treats every parameter equally and cannot pinpoint which parameters actually drive the over-confident behavior. In particular, only a small number of neurons account for most of the over-confident logits, and scaling all parameters uniformly toward the base cannot selectively relax these neurons without weakening the rest. We therefore add a targeted step that first locates the neurons most responsible for the collapsed token distributions via direct logit attribution, a residual-stream decomposition technique commonly used in Transformer circuit analysis~\citep{elhage2021mathematical}, and then resets these neurons in $\theta_{\mathrm{fuse}}$ back to their base-model values.

\subsubsection{Identifying abnormal logit contributors}

\paragraph{Step 1: Selecting over-confident target tokens.}
Over-confidence concentrates on a subset of response tokens rather than being spread uniformly, so we first find where the effect is strongest and then attribute the logits only at those positions. Given a calibration set of prompt--response pairs, we run both the base model and \over in teacher-forcing mode and record the per-position token entropy $H_{\base}(t)$ and $H_{\over}(t)$. For each sample, we pick the Top-$N$ response positions with the largest entropy gap $|H_{\over}(t) - H_{\base}(t)|$ as the target tokens $\mathcal{T}$. These are the positions where \over has most aggressively sharpened the distribution relative to the base model, making them the most informative anchors for attribution.

\paragraph{Step 2: Decomposing logits into per-neuron contributions.}
Under a frozen final-RMSNorm approximation, the target logit on \over can be written as a sum of residual-stream contributions, projected onto a fixed direction that depends on the target token $y^*$:
\begin{equation}
\begin{aligned}
\tilde z_{y^*}(t)
&\;=\;
b_{y^*}
\;+\;
\langle e(t), d_{y^*}(t) \rangle
\;+\;
\sum_{\ell}
\Big[
\langle r^{\mathrm{attn}}_\ell(t), d_{y^*}(t) \rangle
+
\langle r^{\mathrm{mlp}}_\ell(t), d_{y^*}(t) \rangle
\Big], \\
d_{y^*}(t) &\;=\; s_{\mathrm{RMS}}(t) \cdot \big(\gamma \odot W_U[y^*]\big),
\end{aligned}
\label{eq:dla_decomp}
\end{equation}
where $b_{y^*}$ is the optional \texttt{lm\_head} bias for $y^*$\footnote{$b_{y^*}$ denotes the \texttt{lm\_head} bias for token $y^*$ if present (e.g., models without tied embeddings); for models with bias-free \texttt{lm\_head} (as in Llama~\citep{Llama3} and Qwen~\citep{Qwen2.5, Qwen3}), this term vanishes.}, $e(t)$ is the token-embedding write, $r^{\mathrm{attn}}_\ell(t)$ and $r^{\mathrm{mlp}}_\ell(t)$ are the residual-stream writes from the attention and MLP blocks at layer $\ell$, $\gamma$ is the gain of the final RMSNorm, $s_{\mathrm{RMS}}(t)$ is its normalization scalar at position $t$, and $W_U[y^*]$ is the unembedding row of $y^*$. This decomposition becomes exact only because we freeze $s_{\mathrm{RMS}}(t)$ at its forward-pass value and replace the nonlinear RMSNorm with the linear projection $d_{y^*}(t)$. In our implementation we therefore track the reconstruction error $|\tilde z_{y^*}(t) - z_{y^*}(t)|$ to confirm that this approximation is tight. 
We use the resulting scores as attribution signals for ranking rather than as a complete causal explanation, since direct logit attribution can be misleading when later components erase or overwrite earlier residual-stream directions~\citep{janiak2024adversarial}.

For the modules that directly write to the residual stream, namely \texttt{o\_proj} and \texttt{down\_proj}, equation~\eqref{eq:dla_decomp} can be further pushed down to the input-neuron level, giving an exact per-neuron contribution
\begin{equation}
s^{(\ell, m)}_{i} \;=\; a^{(\ell, m)}_{i}(t) \cdot \big\langle W^{(\ell, m)}_{:,i}, \; d_{y^*}(t) \big\rangle,
\qquad m \in \{\texttt{o\_proj},\, \texttt{down\_proj}\},
\label{eq:dla_neuron}
\end{equation}
where $a^{(\ell, m)}_{i}(t)$ is the neuron activation at position $t$ and $W^{(\ell, m)}_{:,i}$ is the corresponding column of the writer weight.

For the remaining internal projections (\texttt{q\_proj}, \texttt{k\_proj}, \texttt{v\_proj}, \texttt{up\_proj}, \texttt{gate\_proj}), whose outputs do not directly write to the residual stream, no such exact neuron-level decomposition is available. We therefore score them with a local ``gradient $\times$ activation'' proxy on the same target logit $z_{y^*}(t)$. This proxy can be viewed as a first-order Taylor approximation to the change in the target logit induced by perturbing or removing the corresponding activation, and has been commonly used as a gradient-based importance estimate for neural units and Transformer components~\citep{molchanov2017pruning, michel2019sixteen}.
Specifically, for \texttt{k\_proj} and \texttt{v\_proj} the score is aggregated over all prefix positions $0..t$, since these projections are read by attention at every later step. For \texttt{q\_proj}, \texttt{up\_proj}, and \texttt{gate\_proj}, it is taken at the target position only. For \texttt{q\_proj} and \texttt{k\_proj}, the implementation uses pre-RoPE projection outputs, so these scores should be read as local attribution proxies rather than exact causal decompositions. This gives a uniform per-neuron score $s^{(\ell, m)}_i$ across all linear modules of the network, which we use for ranking in Section~\ref{sec:analysis_reset}. The visualization of logit attribution is shown in Figure~\ref{fig:logit_attribution} in Appendix~\ref{app:exp_results}.

\subsubsection{Resetting high-attribution neurons}

\paragraph{Building the reset set.}
For each selected target token $t$, we rank all neurons across all replaceable modules by their most positively contributing to the target logit and keep the top $\rho$ fraction, yielding a token-level set $\mathcal{S}_{x,t}$ for each prompt $x$. The final reset set is the union over all selected tokens and calibration examples:
\begin{equation}
  \mathcal{S} = \bigcup_{(x,t)\in\mathcal T} \mathcal S_{x,t}.
\end{equation}
In our implementation, $\rho$ controls the per-token selection budget; we report the resulting effective whole-model reset ratio after taking the union.
We introduce a token-level gating variable $\omega^{(i)} = \mathbb{I}(i \in \mathcal{S}) \in \{0,1\}$ based on the above definition.

\paragraph{Reset operation.}
Inspired by attribution-based neuron intervention in pretrained Transformers~\citep{dai2022knowledge}, we then overwrite each selected neuron in $\theta_{\mathrm{fuse}}$ with its base-model value while leaving the rest untouched:
\begin{equation}
\theta^{(i)}_{\mathrm{reset}} = \omega^{(i)} \theta^{(i)}_{\base} + (1-\omega^{(i)}) \theta^{(i)}_{\mathrm{fuse}}
\label{eq:neuron_reset}
\end{equation}
where each neuron index corresponds to either a row or a column of the underlying weight matrix, depending on the module's role in the residual stream: rows for \texttt{q\_proj}, \texttt{k\_proj}, \texttt{v\_proj}, \texttt{up\_proj}, \texttt{gate\_proj}, and columns for \texttt{o\_proj} and \texttt{down\_proj}.
The full procedure is summarized in Algorithm~\ref{alg:recover}.

\paragraph{Why reset helps.}
Reset and fusion play complementary roles. Fusion smooths the parameter space globally and brings the model into a better-conditioned region, but leaves every direction scaled uniformly, so the neurons that dominate the over-confident logits remain close to their over-trained configuration up to a factor of $\alpha$. The reset step replaces exactly these rigid directions with base-model values, breaking the shortcut that produces collapsed token distributions. Because $|\mathcal{S}|$ is very small, the vast majority of SFT-acquired behavior in $\theta_{\mathrm{fuse}}$ is preserved, while diversity is restored precisely where it was most lost. As shown in Figure~\ref{fig:entropy_recovery}, the output entropy on the affected tokens recovers substantially after the reset, and the resulting model becomes more responsive to subsequent RL optimization.

\vspace{-.1in}
\section{Experiments}
\label{sec:exp}
\vspace{-.1in}

\subsection{Setup}
\label{sec:exp_setup}

\paragraph{Models and Baselines.}
We adopt \fourB~\citep{EvoLM} as the base model for mathematical tasks since it is pre-trained on a controlled corpus without evaluation data contamination, which makes the gain from RL more reliable. For agentic tasks, we use \eightB~\citep{Qwen3}, a strong general-purpose backbone widely used in tool-use scenarios. To isolate the effect of plasticity recovery on the SFT-to-RL handoff, we compare \ourmethod against the following baselines: (1) \mod: a moderately trained SFT checkpoint (epoch=2 for \fourB), which serves as a strong upper-bound reference for the SFT-to-RL handoff; (2) \over: the over-trained SFT checkpoint (epoch=32 for \fourB) that exhibits plasticity loss and is the target for our recovery method; (3) \dft~\citep{DFT}: a representative regularized SFT objective that re-weights the cross-entropy loss with token probabilities to mitigate over-confidence during SFT. Following the same protocol, we report the corresponding +RL results obtained by running the same RL recipe on top of each SFT variant.

\paragraph{Training Details.}

For SFT training, we use LlamaFactory~\citep{LlamaFactory} for \fourB and slime~\citep{slime} for \eightB, both with the AdamW~\citep{Adam} optimizer and learning rate of $3.0 \times 10^{-6}$. To probe behavior under extreme over-training, we train \mod for 2 epochs and \over for 32 epochs without any learning rate schedule or weight decay, so that any difference between the two checkpoints comes purely from training duration.
For RL training, we use verl~\citep{verl} for mathematical tasks and slime~\citep{slime} for agentic tasks. Both stages adopt GRPO~\citep{GRPO} as the base RL algorithm with KL loss. For \fourB on math, we use a learning rate of $1\times 10^{-6}$, sample 8 responses per prompt, and apply rule-based rewards on the boxed final answer. For \eightB on \TB Retail, we follow the official \TB protocol and use a separate Qwen3-4B-Instruct-2507~\citep{Qwen3} as the user simulator.
We provide full hyper-parameter tables and infrastructure details in Appendix~\ref{app:exp_setup}.

\paragraph{Evaluation Details.}

For mathematical reasoning, we evaluate on five widely used in-distribution (ID) benchmarks: GSM8K~\citep{GSM8K}, MATH-500~\citep{PRM800K}, AMC23~\citep{AMC23}, Minerva Math~\citep{Minerva-Math}, and OlympiadBench~\citep{OlympiadBench}. For agentic tasks, we evaluate on \TB~\citep{Tau-Bench} Retail and Airline. We further evaluate on three out-of-distribution (OOD) benchmarks: GPQA-Diamond~\citep{GPQA}, ARC-Challenge~\citep{ARC}, and MMLU-Pro~\citep{MMLU-Pro} following~\citet{LUFFY}. We report Pass@1, which is averaged over multiple samples to ensure robust evaluation. Decoding parameters, per-benchmark $K$, and the full evaluation details are shown in Appendix~\ref{app:exp_setup}.

\subsection{Main Results}
\label{sec:exp_results}

\begin{table*}[!t]
\centering
\small
\caption{Evaluation results on ID and OOD benchmarks of \fourB. The highest values before RL are \underline{underlined} and the highest values after RL are \textbf{bolded}.}
\resizebox{1.0\textwidth}{!}{
\begin{tabular}{
L{1.99cm}   
C{0.99cm}   
C{0.80cm}   
C{0.95cm}   
C{0.75cm}   
C{0.90cm}   
C{0.50cm}   
C{0.80cm}   
C{0.60cm}   
C{0.95cm}   
C{0.55cm}   
C{1.75cm}   
}
\toprule
\multirow{2}{*}{\textbf{Method}} & \multirow{2}{*}{\textbf{GSM8K}} & \multirow{2}{*}{\textbf{MATH}} & \multirow{2}{*}{\textbf{AMC23}} & \multirow{2}{*}{\textbf{Miner.}} & \multirow{2}{*}{\textbf{Olymp.}} & \multirow{2}{*}{\makecell[c]{\textbf{ID}\\\textbf{Avg.}}} & \multirow{2}{*}{\textbf{GPQA}} & \multirow{2}{*}{\textbf{ARC}} & \multirow{2}{*}{\textbf{MMLU}} & \multirow{2}{*}{\makecell[c]{\textbf{OOD}\\\textbf{Avg.}}} & \multirow{2}{*}{\textbf{Avg.}} \\
& \\
\midrule
\mod             & 28.1 & 15.2 & 4.3 & 4.0  & 2.4 & 10.8 & 5.3  & 13.4 & \underline{3.4} & 7.4  & 9.1  \\
\hspace{1em} +RL & \textbf{51.6} & 25.7 & 6.4 & \textbf{11.6} & \textbf{4.7} & \textbf{20.0} & \textbf{13.2} & 22.9 & 6.2 & 14.1 & 17.0 (+7.9) \\
\over            & \underline{31.7} & \underline{15.8} & 3.8 & 5.1  & 2.4 & \underline{11.8} & \underline{7.8}  & \underline{13.7} & 3.3 & \underline{8.3}  & \underline{10.0} \\
\hspace{1em} +RL & 45.2 & 22.0 & 6.6 & 9.9  & 4.1 & 17.5 & 9.9  & 21.3 & 5.6 & 12.2 & 14.9 (+4.9) \\
\dft             & 30.9 & 14.5 & \underline{5.3} & \underline{5.2}  & 2.3 & 11.7 & 5.8  & 7.2  & 3.3 & 5.4  & 8.5  \\
\hspace{1em} +RL & 43.6 & 24.4 & 7.5 & 11.9 & 3.9 & 18.3 & 8.3  & 22.3 & 5.7 & 12.1 & 15.2 (+6.7) \\
\ourmethod & 18.5 & 12.0 & 3.0 & 4.3 & \underline{2.5} & 8.1 & 3.6 & 2.6 & 2.1 & 2.8 & 5.4 \\
\hspace{1em} +RL  & 49.1 & \textbf{26.2} & \textbf{8.0} & 10.8 & 4.2 & 19.7 & 12.9 & \textbf{26.2} & \textbf{6.9} & \textbf{15.3} & \textbf{17.5 (+12.1)} \\
\bottomrule
\end{tabular}%
}
\label{tab:main_math}%
\end{table*}%

\paragraph{Math.}

Table~\ref{tab:main_math} reports the evaluation results on \fourB and we have three observations as follows: (1) Starting RL from \over leads to a clear performance drop compared with \mod (e.g., 17.5 vs. 20.0 ID Avg. and 12.2 vs. 14.1 OOD Avg.), confirming that prolonged SFT actively hurts the subsequent RL stage rather than only saturating it. (2) Replacing SFT with \dft slows down the SFT-induced collapse but does not fully recover the gap, since it modifies the SFT objective in advance and does not address an already over-trained checkpoint. (3) \ourmethod, applied as a post-hoc operation on top of \over, recovers and surpasses the \mod{}+RL upper bound on the average score, with particularly large improvements on the OOD tasks. This indicates that smoothing the parameter landscape and resetting over-confident neurons not only restores RL trainability but also retains the broader knowledge acquired during pre-training, leading to better OOD generalization.

\paragraph{Agentic.}

\begin{table*}[!t]
\centering
\caption{Evaluation results on ID and OOD benchmarks of \eightB. The highest values after RL are \textbf{bolded}.}
\resizebox{0.85\textwidth}{!}{
\begin{tabular}{l*{8}{c}}
\toprule
\multirow{2}{*}{\textbf{Method}} & \multirow{2}{*}{\textbf{\begin{tabular}[c]{@{}c@{}} \TB \\ Retail \end{tabular}}} & \multirow{2}{*}{\textbf{\begin{tabular}[c]{@{}c@{}} \TB \\ Airline \end{tabular}}} & \multirow{2}{*}{\textbf{\begin{tabular}[c]{@{}c@{}} ID \\ Avg. \end{tabular}}} & \multirow{2}{*}{\textbf{GPQA}} & \multirow{2}{*}{\textbf{ARC}} & \multirow{2}{*}{\textbf{MMLU}} & \multirow{2}{*}{\textbf{\begin{tabular}[c]{@{}c@{}} OOD \\ Avg. \end{tabular}}} & \multirow{2}{*}{\textbf{Avg.}} \\
& \\
\midrule
\mod{}+RL & \textbf{78.3} & 16.1 & 47.2 & 40.5 & 77.9 & \textbf{62.5} & 60.3 & 53.8 \\
\over{}+RL & 70.3 & \textbf{19.9} & 45.1 & \textbf{42.0} & 75.9 & 61.7 & 59.9 & 52.5 \\
\dft{}+RL  & 75.3 & 12.2 & 43.8 & 38.8 & 76.8 & \textbf{62.5} & 59.3 & 51.6 \\
\ourmethod{}+RL & 77.0 & 17.5 & \textbf{47.3} & 41.0 & \textbf{78.2} & 62.4 & \textbf{60.5} & \textbf{53.9} \\
\bottomrule
\end{tabular}%
}
\label{tab:main_agent}%
\end{table*}%

Table~\ref{tab:main_agent} summarizes the agentic results on \eightB. Consistent with the math setting, \over{}+RL substantially trails \mod{}+RL on the ID \TB Retail and Airline tasks, where the over-confident policy fails to explore alternative tool-use trajectories and quickly converges to suboptimal behaviors.
\ourmethod lifts the success rate of \over back to (and beyond) the \mod{}+RL reference on both \TB tasks, while simultaneously improving the OOD scores on GPQA, ARC, and MMLU. This shows that the recovered plasticity transfers across rather different task families: the same fusion-and-reset operation that helps reasoning RL also enables tool-use RL to keep learning from environment feedback.

\subsection{Ablation Study}

\paragraph{Components.}
To understand the contribution of each component in \ourmethod, we ablate it on \fourB. As shown in Table~\ref{tab:abla_components}, applying base-anchored fusion alone already brings most of the gain over \over{}+RL: the smoother parameter region restores RL trainability and lifts both ID and OOD scores. Adding the attribution-guided neuron reset on top further improves the average performance, especially on OOD benchmarks. This is consistent with our analysis in Section~\ref{sec:analysis_reset}: fusion shrinks the global SFT-induced drift, while the targeted reset selectively breaks the small set of over-confident logit-contributing directions that fusion alone cannot relax.

\begin{table*}[!h]
\centering
\small
\caption{Evaluation results on ID and OOD benchmarks. The highest values are \textbf{bolded}.}
\resizebox{1.0\textwidth}{!}{
\begin{tabular}{
L{2.85cm}   
C{0.99cm}   
C{0.80cm}   
C{0.95cm}   
C{0.75cm}   
C{0.90cm}   
C{0.50cm}   
C{0.80cm}   
C{0.60cm}   
C{0.95cm}   
C{0.55cm}   
C{0.85cm}   
}
\toprule
\multirow{2}{*}{\textbf{Method}} & \multirow{2}{*}{\textbf{GSM8K}} & \multirow{2}{*}{\textbf{MATH}} & \multirow{2}{*}{\textbf{AMC23}} & \multirow{2}{*}{\textbf{Miner.}} & \multirow{2}{*}{\textbf{Olymp.}} & \multirow{2}{*}{\makecell[c]{\textbf{ID}\\\textbf{Avg.}}} & \multirow{2}{*}{\textbf{GPQA}} & \multirow{2}{*}{\textbf{ARC}} & \multirow{2}{*}{\textbf{MMLU}} & \multirow{2}{*}{\makecell[c]{\textbf{OOD}\\\textbf{Avg.}}} & \multirow{2}{*}{\textbf{Avg.}} \\
& \\
\midrule
\over & 45.2 & 22.0 & 6.6 & 9.9  & 4.1 & 17.5 & 9.9  & 21.3 & 5.6 & 12.2 & 14.9 \\
\over w/Fusion & 46.9 & \textbf{26.6} & 6.3 & \textbf{12.2} & \textbf{4.3} & 19.3 & \textbf{14.7} & 22.7 & \textbf{7.0} & 14.8 & 17.0 \\
\ourmethod & \textbf{49.1} & 26.2 & \textbf{8.0} & 10.8 & 4.2 & \textbf{19.7} & 12.9 & \textbf{26.2} & 6.9 & \textbf{15.3} & \textbf{17.5} \\
\bottomrule
\end{tabular}%
}
\label{tab:abla_components}%
\end{table*}%

\paragraph{Fusion Weight.}

We further sweep the fusion weight $\alpha$ in $\{0.4, 0.45, 0.5, 0.55, 0.6\}$ on \fourB to understand how strongly the over-trained checkpoint should be pulled toward the base. As shown in Table~\ref{tab:abla_fusion_weight}, a small $\alpha$ (e.g., 0.4) drops more of the SFT-acquired ID skills but leaves the model in a smoother region with stronger OOD learning potential, whereas a large $\alpha$ (e.g., 0.6) preserves more SFT behavior at the cost of carrying over the rigidity in both ID and OOD evaluation. A moderate value of $\alpha=0.5$ balances the two effects best on average and is used as the default in all main experiments.

\begin{table*}[!h]
\centering
\small
\caption{Evaluation results on ID and OOD benchmarks with different fusion weight $\alpha$. The highest values are \textbf{bolded}.}
\resizebox{0.95\textwidth}{!}{
\begin{tabular}{
L{1.20cm}   
C{0.99cm}   
C{0.80cm}   
C{0.95cm}   
C{0.75cm}   
C{0.90cm}   
C{0.50cm}   
C{0.80cm}   
C{0.60cm}   
C{0.95cm}   
C{0.55cm}   
C{0.85cm}   
}
\toprule
\multirow{2}{*}{\textbf{Method}} & \multirow{2}{*}{\textbf{GSM8K}} & \multirow{2}{*}{\textbf{MATH}} & \multirow{2}{*}{\textbf{AMC23}} & \multirow{2}{*}{\textbf{Miner.}} & \multirow{2}{*}{\textbf{Olymp.}} & \multirow{2}{*}{\makecell[c]{\textbf{ID}\\\textbf{Avg.}}} & \multirow{2}{*}{\textbf{GPQA}} & \multirow{2}{*}{\textbf{ARC}} & \multirow{2}{*}{\textbf{MMLU}} & \multirow{2}{*}{\makecell[c]{\textbf{OOD}\\\textbf{Avg.}}} & \multirow{2}{*}{\textbf{Avg.}} \\
& \\
\midrule
0.4  & 47.7 & 23.9 & 6.2 & 10.3 & 4.5 & 18.5 & 12.8 & \textbf{26.4} & 5.9 & \textbf{15.0} & 16.8 \\
0.45 & 46.6 & 26.4 & 6.2 & 10.6 & 4.0 & 18.8 & \textbf{14.7} & 23.3 & 6.4 & 14.8 & 16.8 \\
0.5 & 46.9 & 26.6 & 6.3 & \textbf{12.2} & 4.3 & 19.3 & \textbf{14.7} & 22.7 & \textbf{7.0} & 14.8 & \textbf{17.0} \\
0.55 & 48.8 & \textbf{26.8} & \textbf{9.5} & 12.0 & 3.5 & \textbf{20.1} & 9.3 & 23.2 & 6.6 & 13.0 & 16.6 \\
0.6 & \textbf{50.1} & \textbf{26.8} & 6.8 & 9.8 & \textbf{5.0} & 19.7 & 8.8 & 21.0 & 5.9 & 11.9 & 15.8 \\
\bottomrule
\end{tabular}%
}
\vspace{-1em}
\label{tab:abla_fusion_weight}%
\end{table*}%

\paragraph{Reset Percentage.}

We then study the effect of the per-token reset budget $\rho$, which controls how many of the most positively contributing neurons are rolled back to their base-model values per target token. We sweep $\rho \in \{0.5\%, 1\%, 2\%, 4\%\}$. As shown in Table~\ref{tab:abla_reset_percentage}, even a very small $\rho$ already yields a noticeable improvement, indicating that the over-confident behavior concentrates in a tiny fraction of neurons. Moderate values further improve OOD generalization, while overly aggressive reset starts to erase useful SFT-acquired behaviors and hurts ID accuracy. We therefore adopt a small default reset ratio for all main experiments.

\begin{table*}[!h]
\centering
\small
\caption{Evaluation results on ID and OOD benchmarks with different reset percentage $\rho$. The highest values are \textbf{bolded}.}
\resizebox{0.95\textwidth}{!}{
\begin{tabular}{
L{1.20cm}   
C{0.99cm}   
C{0.80cm}   
C{0.95cm}   
C{0.75cm}   
C{0.90cm}   
C{0.50cm}   
C{0.80cm}   
C{0.60cm}   
C{0.95cm}   
C{0.55cm}   
C{0.85cm}   
}
\toprule
\multirow{2}{*}{\textbf{Method}} & \multirow{2}{*}{\textbf{GSM8K}} & \multirow{2}{*}{\textbf{MATH}} & \multirow{2}{*}{\textbf{AMC23}} & \multirow{2}{*}{\textbf{Miner.}} & \multirow{2}{*}{\textbf{Olymp.}} & \multirow{2}{*}{\makecell[c]{\textbf{ID}\\\textbf{Avg.}}} & \multirow{2}{*}{\textbf{GPQA}} & \multirow{2}{*}{\textbf{ARC}} & \multirow{2}{*}{\textbf{MMLU}} & \multirow{2}{*}{\makecell[c]{\textbf{OOD}\\\textbf{Avg.}}} & \multirow{2}{*}{\textbf{Avg.}} \\
& \\
\midrule
0.5\% & 49.1 & 25.9 & 10.7 & 12.0 & 3.9 & 20.3 & 11.6 & 21.3 & 6.0 & 13.0 & 16.7 \\
1\%  & \textbf{49.1} & 26.2 & \textbf{8.0} & 10.8 & 4.2 & 19.7 & 12.9 & \textbf{26.2} & 6.9 & \textbf{15.3} & \textbf{17.5} \\
2\% & 48.9 & 25.5 & 7.3 & \textbf{12.2} & \textbf{5.4} & \textbf{19.9} & \textbf{13.7} & 23.4 & \textbf{7.3} & 14.8 & 17.3 \\
4\% & 46.8 & 24.7 & 5.0 & 10.1 & 4.8 & 18.3 & 9.9 & 22.6 & 6.6 & 13.0 & 15.6 \\
\bottomrule
\end{tabular}%
}
\vspace{-1.1em}
\label{tab:abla_reset_percentage}%
\end{table*}%

\paragraph{\ourmethod with different SFT checkpoints.}
Table~\ref{tab:abla_sft_model} further studies whether \ourmethod is specific to severely over-trained checkpoints or can also
be applied to checkpoints with different SFT degrees.
Starting RL from \mod already gives strong ID performance, but applying \ourmethod before RL
further improves the overall average score from 17.0 to 17.6, mainly by increasing the OOD average from 14.1 to 16.3.
This suggests that even a moderately trained SFT checkpoint may still contain SFT-induced rigidity, and a mild recovery
operation can improve its ability to generalize after RL.
However, this improvement comes with a small drop on the ID average, indicating a trade-off between preserving task-specific
SFT behavior and restoring broader plasticity.
These results indicate that the proposed recovery procedure is not merely an early-stopping substitute: it can rejuvenate
checkpoints from different stages of SFT.
  
\begin{table*}[!h]
\centering
\small
\caption{Evaluation results on ID and OOD benchmarks of different methods after RL training. The highest values are \textbf{bolded}.}
\resizebox{1.0\textwidth}{!}{
\begin{tabular}{
L{3.39cm}   
C{0.99cm}   
C{0.80cm}   
C{0.95cm}   
C{0.75cm}   
C{0.90cm}   
C{0.50cm}   
C{0.80cm}   
C{0.60cm}   
C{0.95cm}   
C{0.55cm}   
C{0.85cm}   
}
\toprule
\multirow{2}{*}{\textbf{Method}} & \multirow{2}{*}{\textbf{GSM8K}} & \multirow{2}{*}{\textbf{MATH}} & \multirow{2}{*}{\textbf{AMC23}} & \multirow{2}{*}{\textbf{Miner.}} & \multirow{2}{*}{\textbf{Olymp.}} & \multirow{2}{*}{\makecell[c]{\textbf{ID}\\\textbf{Avg.}}} & \multirow{2}{*}{\textbf{GPQA}} & \multirow{2}{*}{\textbf{ARC}} & \multirow{2}{*}{\textbf{MMLU}} & \multirow{2}{*}{\makecell[c]{\textbf{OOD}\\\textbf{Avg.}}} & \multirow{2}{*}{\textbf{Avg.}} \\
& \\
\midrule
\mod{} & \textbf{51.6} & 25.7 & 6.4 & \textbf{11.6} & \textbf{4.7} & \textbf{20.0} & 13.2 & 22.9 & 6.2 & 14.1 & 17.0 \\
\mod{}+\ourmethod{} & 46.9 & 24.3 & \textbf{8.2} & 10.8 & 4.5 & 18.9 & \textbf{14.2} & \textbf{26.3} & \textbf{8.4} & \textbf{16.3} & \textbf{17.6} \\
\over{}+\ourmethod{} & 48.6 & \textbf{26.2} & 8.0 & 10.8 & 4.2 & 19.6 & 12.9 & 26.2 & 6.9 & 15.3 & 17.5 \\
\bottomrule
\end{tabular}%
}
\label{tab:abla_sft_model}%
\end{table*}%

\section{Conclusion}
\label{sec:conclusion}

In this paper, we identify a failure mode where excessive SFT training leads to loss of plasticity, which limits the effectiveness of subsequent RL optimization. Through analysis of parameter changes, token distributions, and RL training dynamics, we show that over-trained models become over-confident and harder to update. To address this issue, we introduce a two-stage method that includes global model fusion and neuron reset. 
These components help smooth the parameter space and restore diversity in key parts of the model, making it more amenable to further optimization during RL.
Empirical results on mathematical and agentic tasks demonstrate the effectiveness of our method. Our approach consistently improves RL performance over over-trained SFT models and also shows better generalization on out-of-distribution benchmarks.
Our findings highlight the importance of model plasticity in the SFT-then-RL pipeline. We hope this work can motivate future research on understanding optimization dynamics in post-training and developing more robust training strategies.

Although our method effectively restores the plasticity of the over-trained SFT models, both model fusion and neuron reset require the access to the base model for reference. We will explore how to restore the plasticity of over-trained models without the base model in the future work.

\clearpage
\bibliography{main}
\bibliographystyle{plainnat}

\clearpage
\appendix

\section{Full Algorithm}

\begin{algorithm}[!h]
\caption{\ourmethod: Plasticity Recovery for Over-Trained SFT Models}
\label{alg:recover}
\begin{algorithmic}[1]
\Require Base model $\theta_{\base}$, over-trained model $\theta_{\mathrm{over}}$, fusion weight $\alpha$, reset ratio $\rho$
\State Apply global model fusion via~\eqref{eq:model_fusion}
\State Compute DLA scores for neurons using over-confident tokens
\State For each selected target token, select the global top-$\rho$ neurons by most positively contributing attribution score
\State Let $\mathcal S$ be the union of selected neurons over all calibration tokens
\State Reset selected neurons in $\theta_{\mathrm{fuse}}$ to their base values
\Ensure recovered model $\theta_{\mathrm{rec}}$
\end{algorithmic}
\end{algorithm}

\section{Experimental Details}
\label{app:exp_setup}

\paragraph{Mathematical SFT Training Details.}

For \fourB, we conduct SFT with LlamaFactory~\citep{LlamaFactory} on a 100k subset sampled from a 500k mathematical SFT corpus. The base checkpoint is the \fourB pre-trained model trained with controlled mixing of FineWeb and FineMath data (without contamination from the evaluation benchmarks). All variants share the same data, the same context window, and the same data ordering, so any difference between \mod and \over is purely a function of training duration.
We use AdamW~\citep{Adam} with $\beta_1=0.9$, $\beta_2=0.999$, a constant learning rate of $3.0\times 10^{-6}$, and \emph{no} learning-rate warmup, decay, or weight decay. We deliberately disable schedules and weight decay so that excessive SFT can manifest its full effect on parameter drift, weight magnitude, and token entropy. \mod is the checkpoint after 2 epochs and \over is the checkpoint after 32 epochs over the same SFT subset.
For the \dft baseline, we keep the same data, the same total epochs (32), and the same optimizer settings as \over, and only replace the cross-entropy loss with the probability-reweighted DFT objective~\citep{DFT}. This isolates the effect of the loss formulation from the effect of the SFT trajectory length.

\paragraph{Agentic SFT Training Details.}

For \eightB, the SFT stage is performed with slime~\citep{slime} on the \TB Retail training split, where supervision trajectories are generated from a stronger Qwen3-30B-A3B teacher and saved as multi-turn assistant-tool conversations. We use AdamW with $\beta_1=0.9$, $\beta_2=0.999$, a constant learning rate of $3.0\times 10^{-6}$, weight decay 0.1, a global batch size of 16, and a maximum response length of 4{,}096 tokens. We train for 32 epochs with the Qwen3 chat template (\texttt{enable\_thinking}=False) and the Qwen3 tool-aware loss mask, so that loss is only computed on assistant turns. The corresponding \mod and \over checkpoints follow the same epoch-based convention as in the math setting. The \dft baseline is constructed in the same way as in the math setting, by replacing the cross-entropy loss with the DFT objective while keeping all other hyper-parameters identical.

\paragraph{Mathematical RL Training Details.}

We use verl~\citep{verl} with GRPO~\citep{GRPO} as the RL backbone. The actor is initialized from the corresponding SFT checkpoint, and the reference policy is the same SFT model frozen at step~0. The training prompt batch size is 512 with a mini-batch size of 128, and we sample 8 responses per prompt with temperature 1.0 and top-$p$ 1.0. The maximum prompt length is 512 and the maximum response length is 1{,}024 tokens. For optimization, we use AdamW with a constant learning rate of $1\times 10^{-6}$ and a global gradient clip of 1.0. We keep the GRPO clipping range symmetric ($\varepsilon_\text{low}=\varepsilon_\text{high}=0.2$), use KL loss with coefficient 0.001, and adopt the \texttt{seq-mean-token-mean} loss aggregation. Rewards are computed by the Math-Verify\footnote{\url{https://github.com/huggingface/Math-Verify}} verifier on the boxed final answer, with the reward function returning 1 for correct answers and -1 otherwise. Validation is run every 50 steps with temperature 0.6 and top-$p$ 1.0. Each run uses 1 node of 8$\times$NVIDIA H800 GPUs, FSDP2 sharding, and dynamic batching. Rollouts are served by vLLM~\citep{vLLM}.

\paragraph{Agentic RL Training Details.}

For agentic RL, we use slime~\citep{slime} with Megatron-LM~\citep{Megatron-LM} backend for training and SGLang~\citep{SGLang} for inference. For each iteration we sample 12 prompts from the \TB Retail training set with 8 trajectories per prompt and a global batch size of 96, with maximum response length 1{,}024 tokens and temperature 1.0. To produce realistic multi-turn behavior, we deploy a separate Qwen3-4B-Instruct-2507~\citep{Qwen3} as the user simulator, served locally via SGLang, and call it through an OpenAI-compatible API. We use AdamW with learning rate $1\times 10^{-6}$, constant LR schedule, and CPU optimizer offloading. We use GRPO algorithm and set the KL loss coefficient and the entropy coefficient both to 0, and use clipping ranges with $\varepsilon_\text{low}=0.2$ and $\varepsilon_\text{high}=0.28$. Tensor-model-parallel size is 2, sequence parallelism is enabled, and we use full activation recomputation with dynamic batching (up to 2{,}048 tokens per GPU). We train for up to 200 rollouts, save and evaluate every 20 rollouts, and balance trajectories across data-parallel ranks with the standard non-zero-reward-std dynamic filter. Each run uses 6 GPUs for the policy and the user simulator combined, with the user simulator pinned to a separate set of GPUs.

For both math and agentic RL, the only difference between baselines and \ourmethod is the \emph{initialization} of the actor: baselines start from \mod, \over, or the \dft checkpoint, whereas \ourmethod starts from the rejuvenated checkpoint produced by Algorithm~\ref{alg:recover}. All RL hyper-parameters, data, prompts, and seeds are kept identical across initializations to ensure a fair comparison.

\paragraph{Evaluation Details.}

For ID and OOD evaluation we use the same inference backends as in training: vLLM for \fourB and SGLang for \eightB. We use a sampling temperature of 0.6 and top-$p$ of 1.0 for all benchmarks. For benchmarks with limited problem counts, we average Pass@1 over multiple samples per problem ($K{=}32$ for AMC23 and AIME-style sets, $K{=}4$ for MATH-500, Minerva, OlympiadBench, GPQA, ARC, $K{=}2$ for GSM8K). Mathematical answers are graded with the same Math-Verify verifier used during RL training, and multiple-choice OOD benchmarks (GPQA, ARC, MMLU-Pro) are graded by the official answer-matching scripts shipped with each dataset.

\section{Additional Results}
\label{app:exp_results}

\paragraph{Logit Distribution.}

\begin{figure*}[!ht]
\centering
\begin{tabular}{cc}
\hspace{-1.1em}
\subfloat[\centering \mod]{\centering\includegraphics[width=0.4\linewidth]{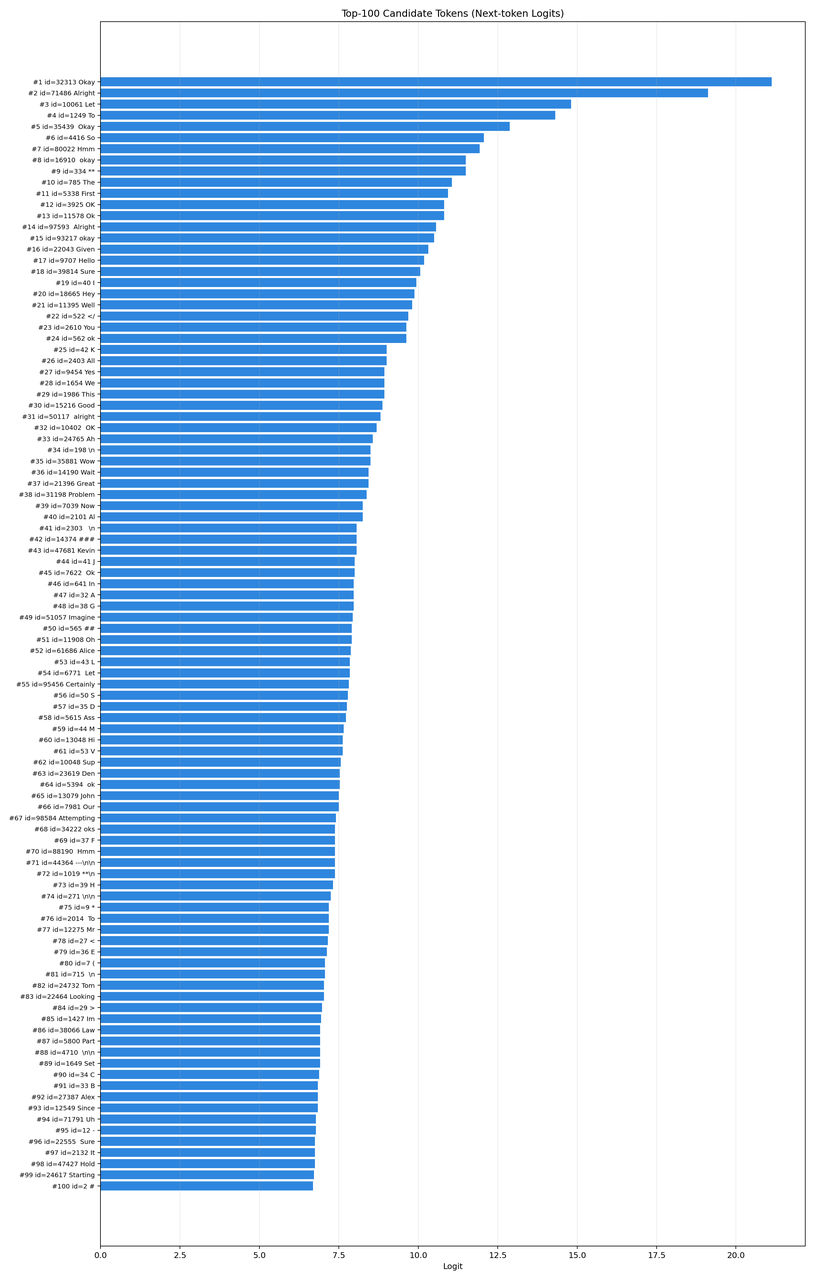}}\label{fig:logit_dist_mod}
& \subfloat[\centering \over]{\includegraphics[width=0.4\linewidth]{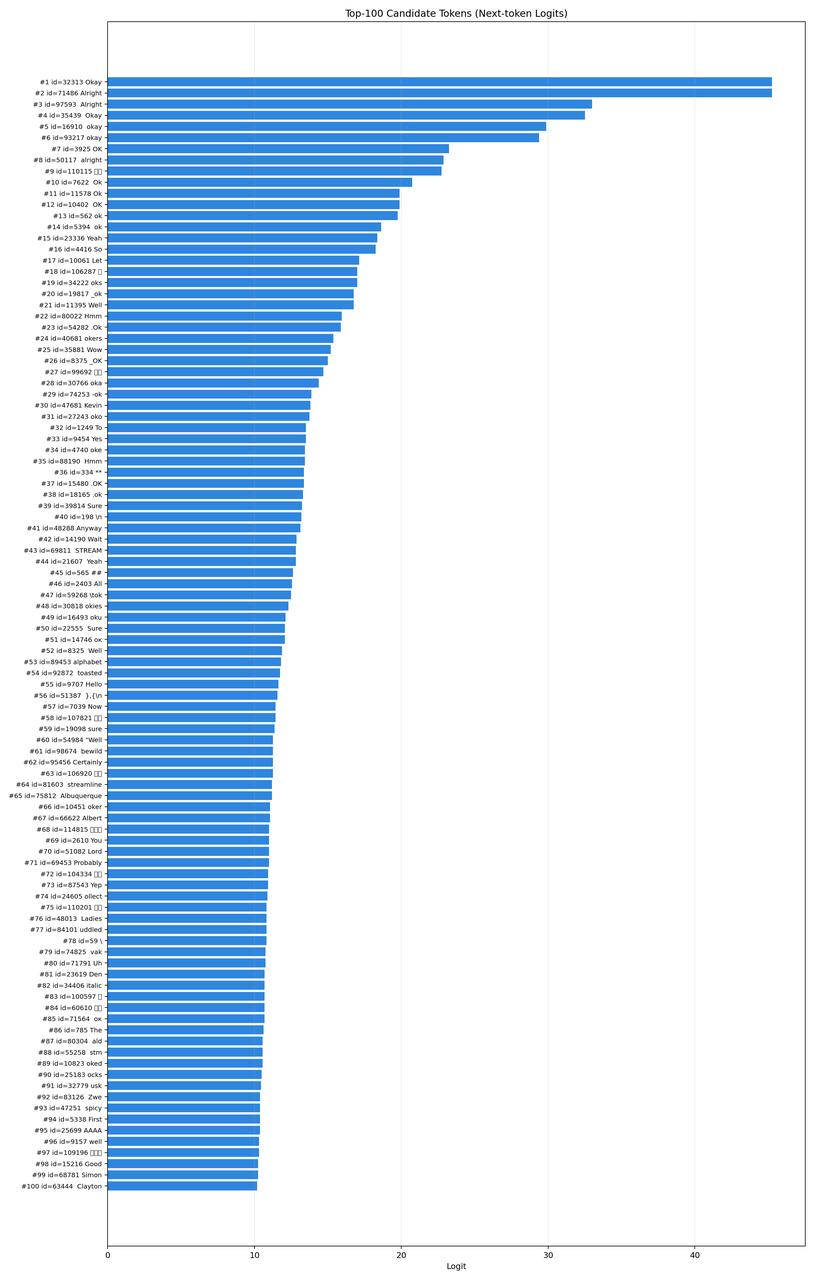}}\label{fig:logit_dist_over}
\end{tabular}
\caption{Token logit comparison between normal SFT models and over-trained models: (a) \mod; (b) \over.}
\label{fig:logit_dist}
\end{figure*}

Figure~\ref{fig:logit_dist} contrasts the next-token logit distribution between \mod and \over on the same set of held-out prompts. Under teacher forcing on the gold response, \over concentrates almost the entire probability mass on a single token, while \mod still allocates non-trivial mass to plausible alternatives. We additionally inspect the per-position entropy along full responses and observe that the entropy gap between the two checkpoints is largest on tokens that involve numerical answers, formula formatting, and chain-of-thought connectors, which is consistent with our DLA-based attribution analysis.

\begin{figure*}[!ht]
\centering
\begin{tabular}{cc}
\hspace{-1.1em}
\subfloat[\centering module level]{\centering\includegraphics[width=0.48\linewidth]{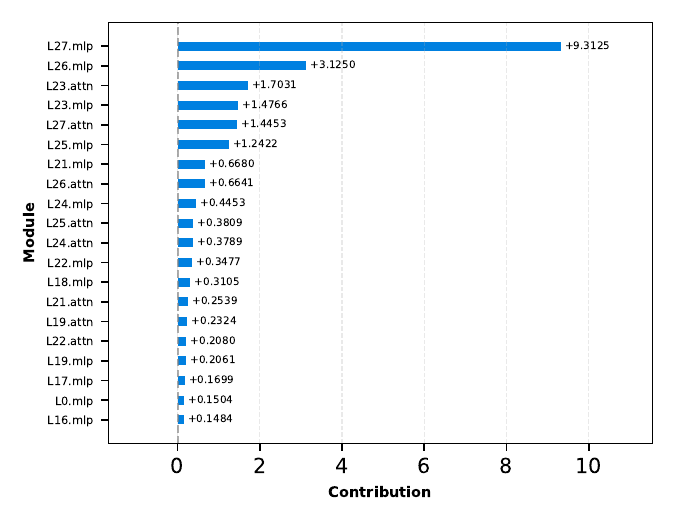}}\label{fig:logit_attribution_module}
\hspace{-1.1em}
& \subfloat[\centering neuron level]{\includegraphics[width=0.48\linewidth]{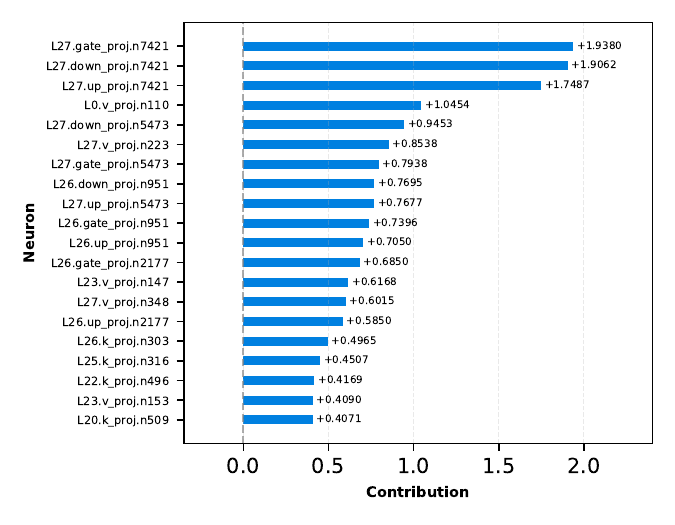}}\label{fig:logit_attribution_neuron}
\end{tabular}
\caption{Logit attribution ranking: (a) module level; (b) neuron level.}
\label{fig:logit_attribution}
\end{figure*}

\paragraph{Logit Attribution.}
The attribution scores are highly uneven across both modules and neurons.
At the module level, only a small number of modules contribute disproportionately large positive values to the over-confident
target logits, while most modules have much smaller contributions, as shown in Figure~\ref{fig:logit_attribution}.
This indicates that the excessive logit sharpening induced by over-training is not uniformly distributed across the network,
but is concentrated in a limited set of components.
A similar pattern appears at the neuron level: within the high-contribution modules, only a small subset of neurons dominates
the positive contribution to the gold-token logit, whereas the majority of neurons contribute little or even negatively.
This heavy-tailed attribution structure explains why resetting a small fraction of high-scoring neurons is sufficient to relax
the over-confident logits and recover plasticity, without broadly disrupting the SFT-acquired behavior stored in the rest of
the model.

\paragraph{Selected Tokens in Logit Attribution.}

\begin{figure*}[!ht]
\centering
\includegraphics[width=0.6\textwidth]{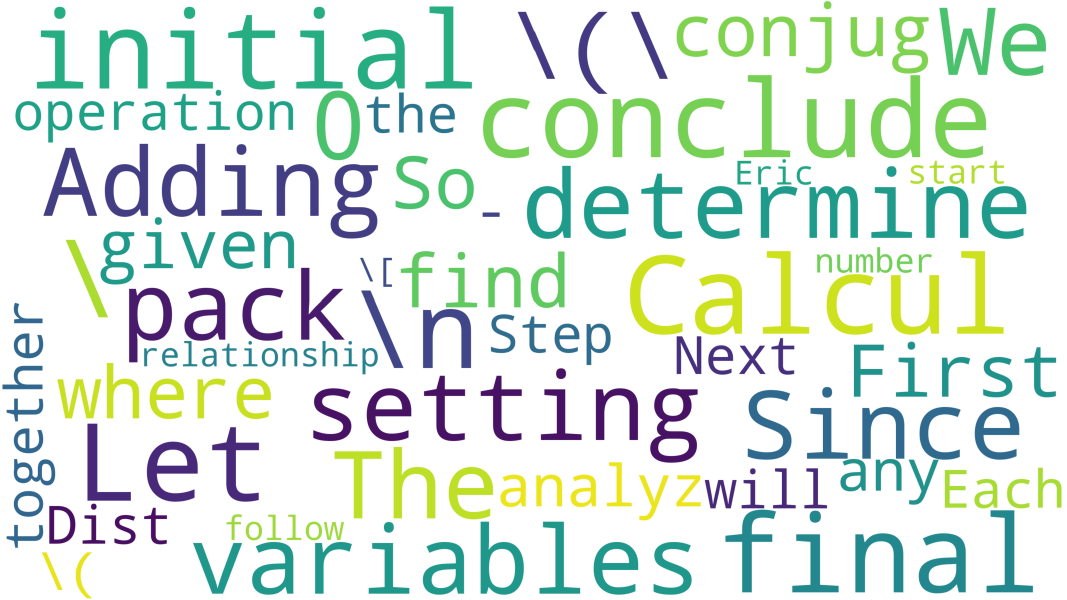}
\caption{Wordcloud of selected tokens in logit attribution.}
\label{fig:wordcloud}
\end{figure*}

We visualize the selected tokens of logit attribution in Figure~\ref{fig:wordcloud}. We can observe that the selected tokens are mainly logical connectives instead of basic knowledge or computation process. As pointed out by previous work~\citep{80/20}, RL mainly optimizes tokens related to reasoning behaviors (e.g., logical connectives). \ourmethod recovers the entropy of these critical tokens, which better incentivizes the learnability of the policy during RL. 

\paragraph{How token entropy recovers?}

\begin{figure*}[!ht]
\centering
\begin{tabular}{cc}
\hspace{-1.1em}
\subfloat[\centering Before neuron reset]{\centering\includegraphics[width=0.48\linewidth]{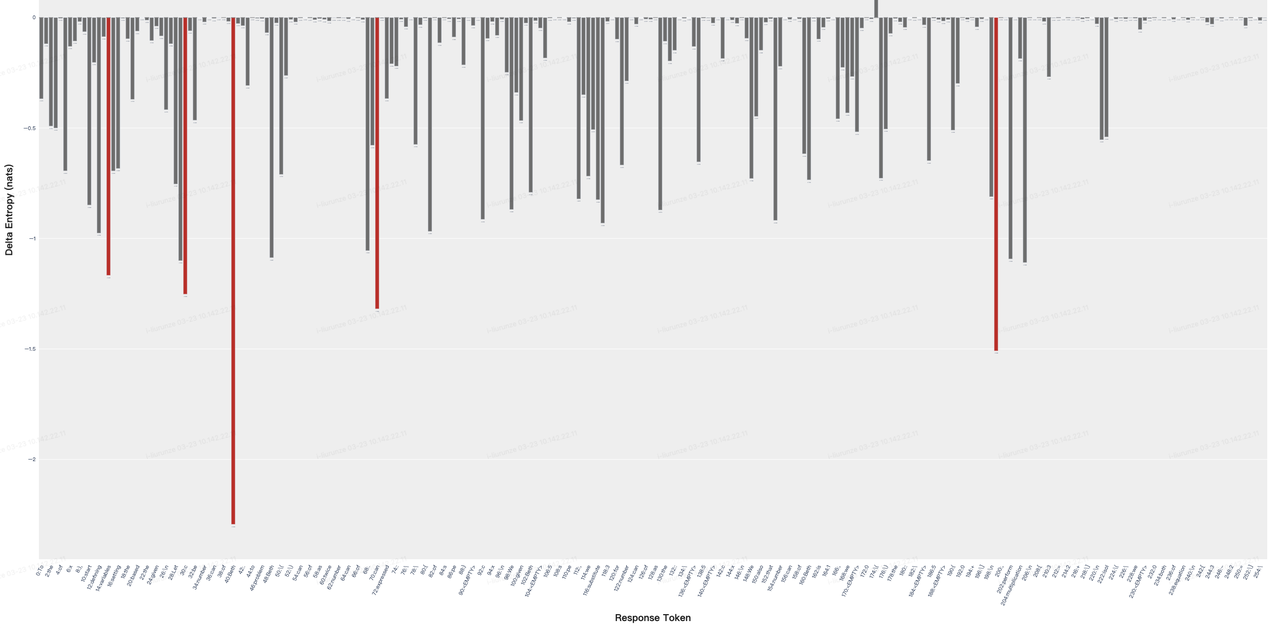}}\label{fig:entropy_before}
& \subfloat[\centering After neuron reset]{\includegraphics[width=0.48\linewidth]{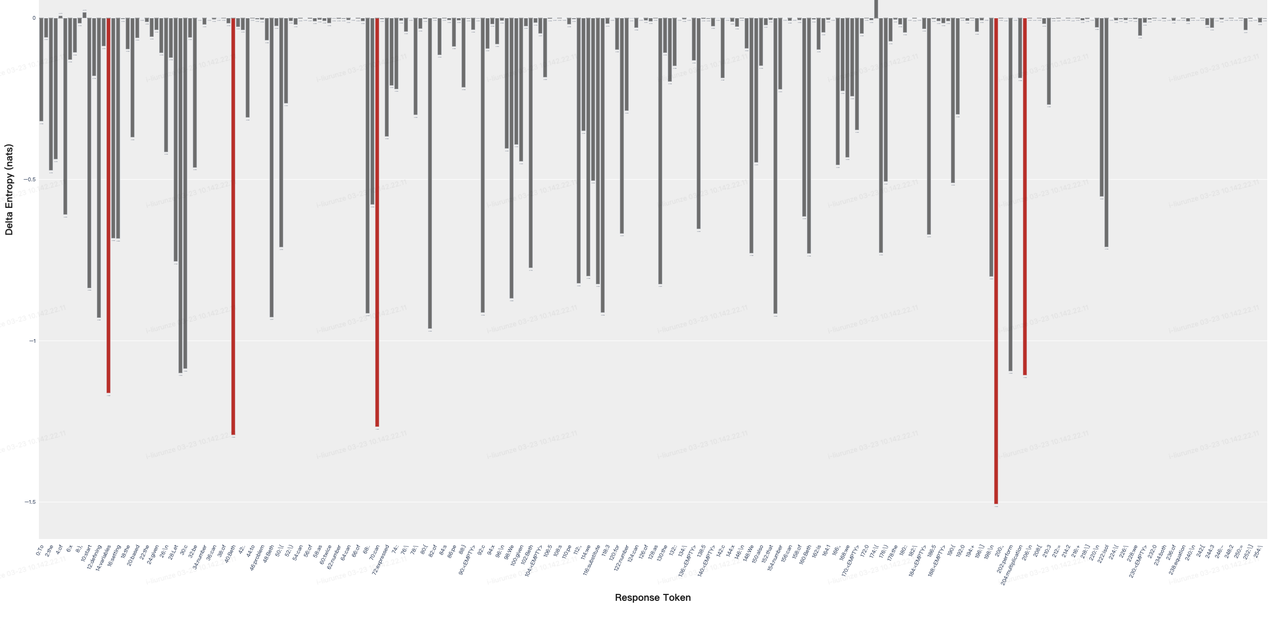}}\label{fig:entropy_after}
\end{tabular}
\caption{Comparison of token entropy: (a) Before neuron reset; (b) After neuron reset.}
\label{fig:entropy_recovery}
\end{figure*}

To better understand how neuron reset restores output diversity, we visualize the token-level entropy before and after applying
the reset operation in Figure~\ref{fig:entropy_recovery}.
Before reset, \over produces extremely low entropy on many selected target positions, indicating that the policy assigns most
probability mass to a single token and leaves little room for alternative reasoning continuations.
After resetting the high-attribution neurons to their base-model values, the entropy of these positions increases noticeably,
while the distribution does not become uniformly random.
This suggests that the reset operation does not simply inject noise into the model. Instead, it selectively relaxes the over-confident logits associated with a small set of abnormal neurons.
As a result, the recovered checkpoint preserves the main SFT-acquired behavior but reopens alternative token choices that are
useful for exploration during RL.
This entropy recovery provides an output-space explanation for why \ourmethod improves subsequent RL optimization and OOD
generalization.

\paragraph{Selected neurons/modules.}

\begin{figure*}[!ht]
\centering
\includegraphics[width=0.8\textwidth]{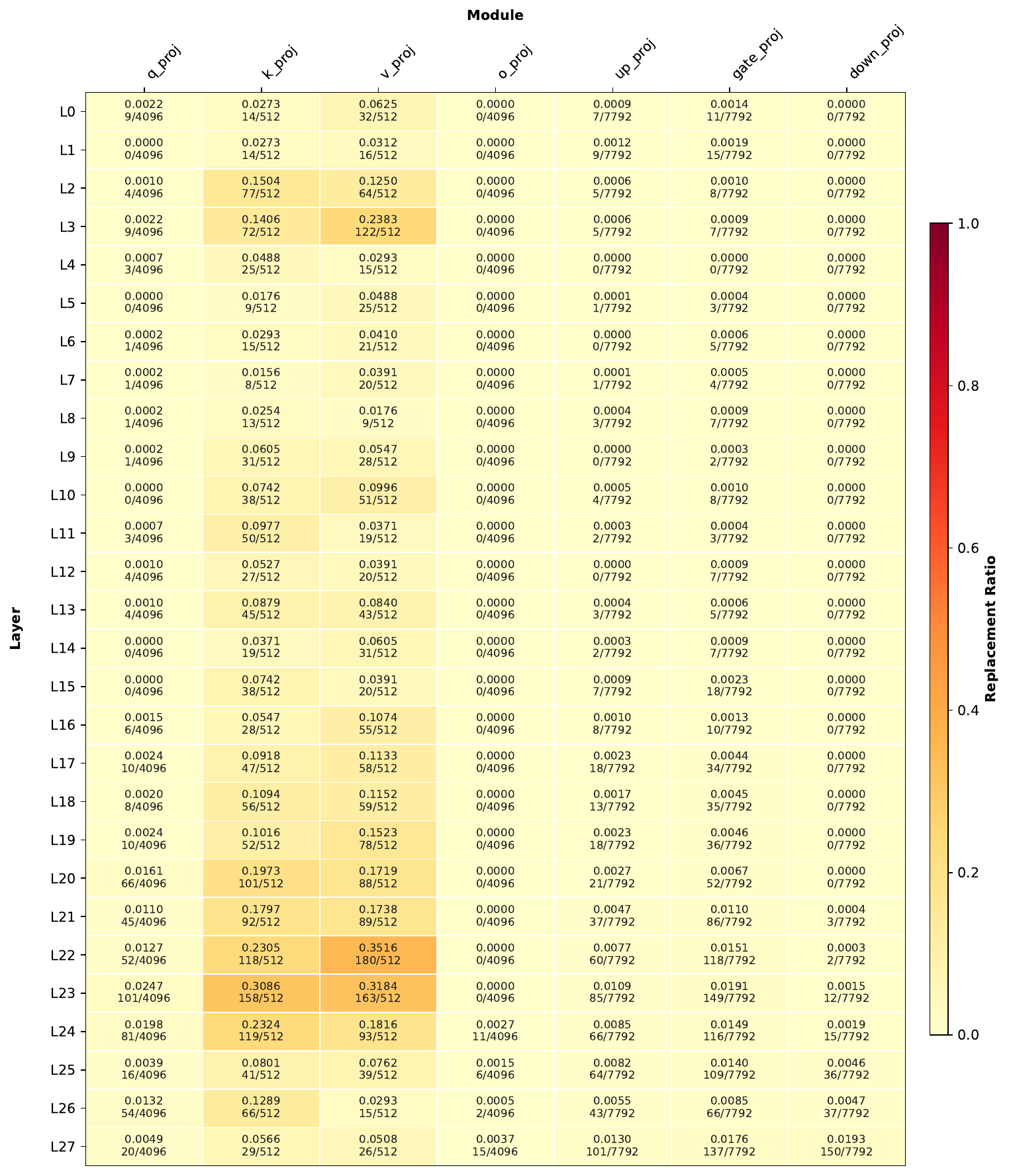}
\caption{Per layer and per module visualization of neuron reset ratio in \fourB.}
\label{fig:replacement_ratio_heatmap}
\end{figure*}

Figure~\ref{fig:replacement_ratio_heatmap} visualizes the distribution of reset neurons over the whole model. Reset neurons are not uniformly distributed: they cluster in a few specific layers and modules that are most aligned with over-confident logit production, especially the \texttt{k\_proj} and \texttt{v\_proj} modules in the last several layers. We also observe that reset neurons aggregated from different calibration prompts overlap substantially, indicating that the over-confidence of \over is governed by a stable, prompt-agnostic subset of parameters rather than per-prompt artifacts. This stability is what makes the reset operation effective with a single calibration pass.

\paragraph{Computational Cost.}

\ourmethod is a one-shot, post-hoc operation and introduces negligible overhead compared with the SFT or RL stages. The fusion step is a single element-wise interpolation between two checkpoints. The neuron reset step requires one forward pass over a small calibration set to compute DLA scores and a single masked copy from the base weights into the fused checkpoint. Both steps are run on a single NVIDIA H800 GPU and finish in 3 minutes for \fourB and within 5 minutes for \eightB, which is orders of magnitude smaller than the cost of either re-running SFT from a different stopping point or relaunching RL from multiple SFT checkpoints.

\paragraph{Parameter Changes.}

\begin{figure*}[!ht]
\centering
\includegraphics[width=0.8\textwidth]{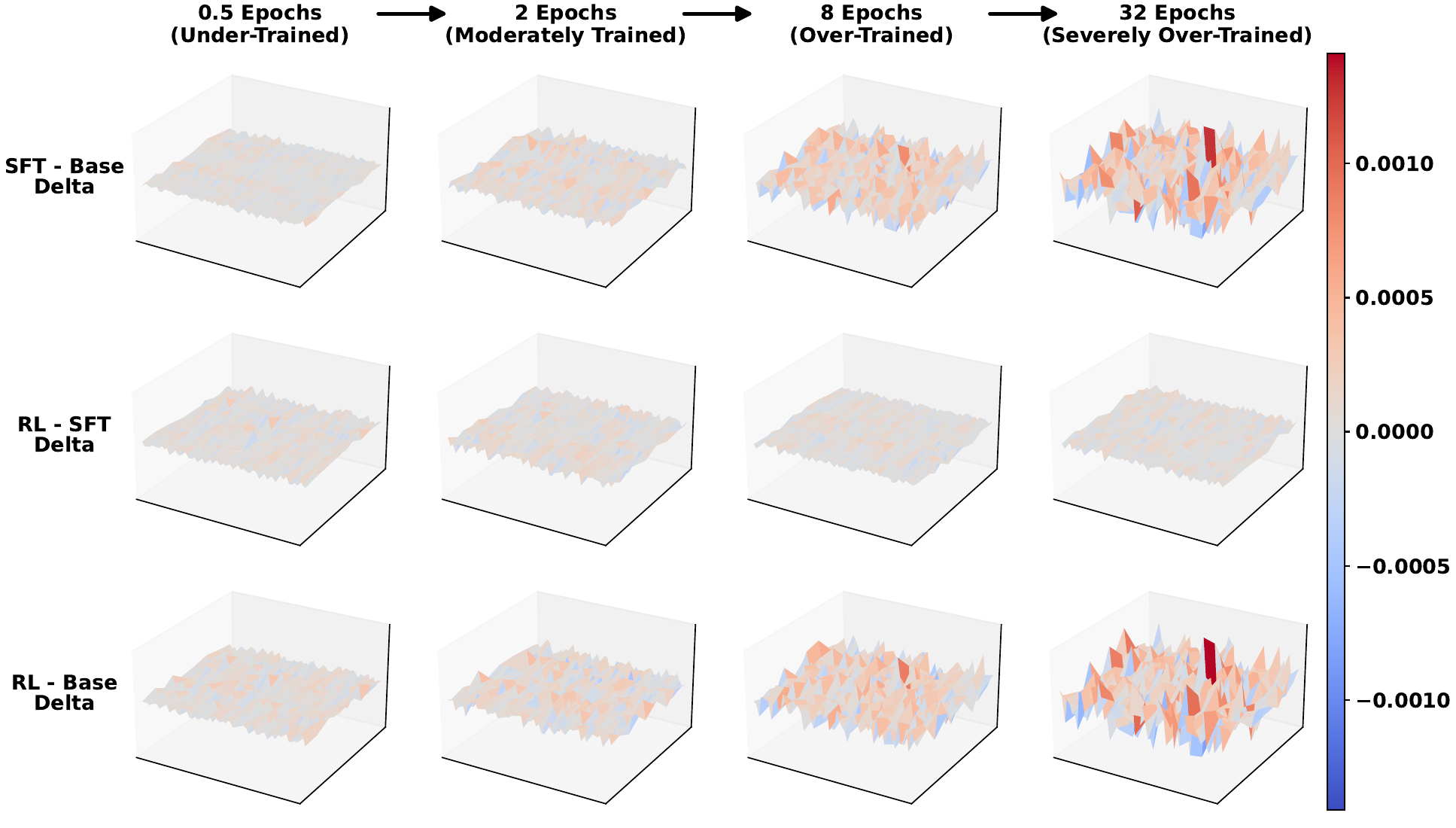}
\caption{Parameter changes of \texttt{layers.0.self\_attn.k\_proj.weight} induced by SFT and RL.}
\label{fig:param_change_l0_kproj}
\end{figure*}

\begin{figure*}[!ht]
\centering
\includegraphics[width=0.8\textwidth]{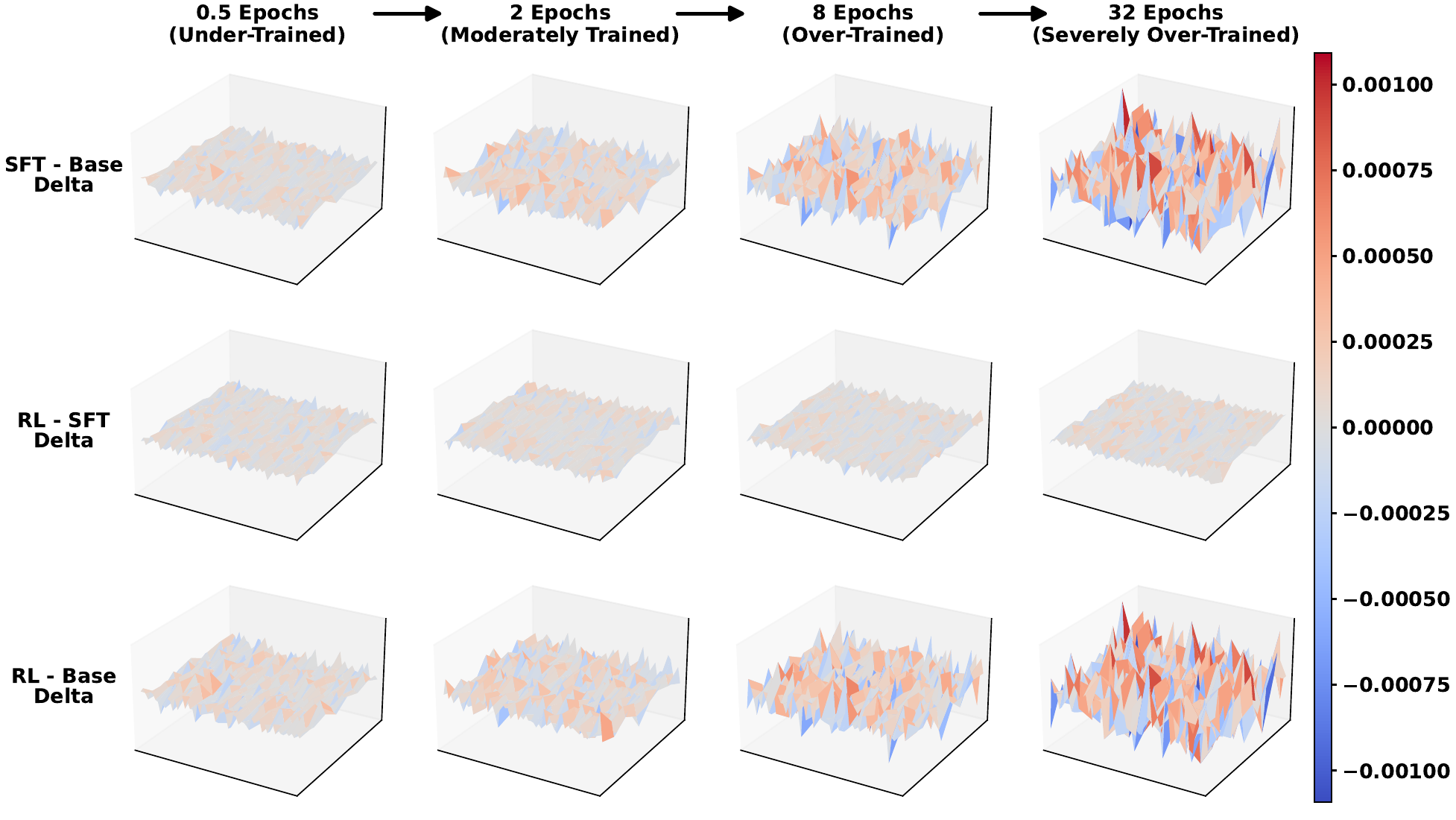}
\caption{Parameter changes of \texttt{layers.27.self\_attn.v\_proj.weight} induced by SFT and RL.}
\label{fig:param_change_l27_vproj}
\end{figure*}

\begin{figure*}[!ht]
\centering
\includegraphics[width=0.8\textwidth]{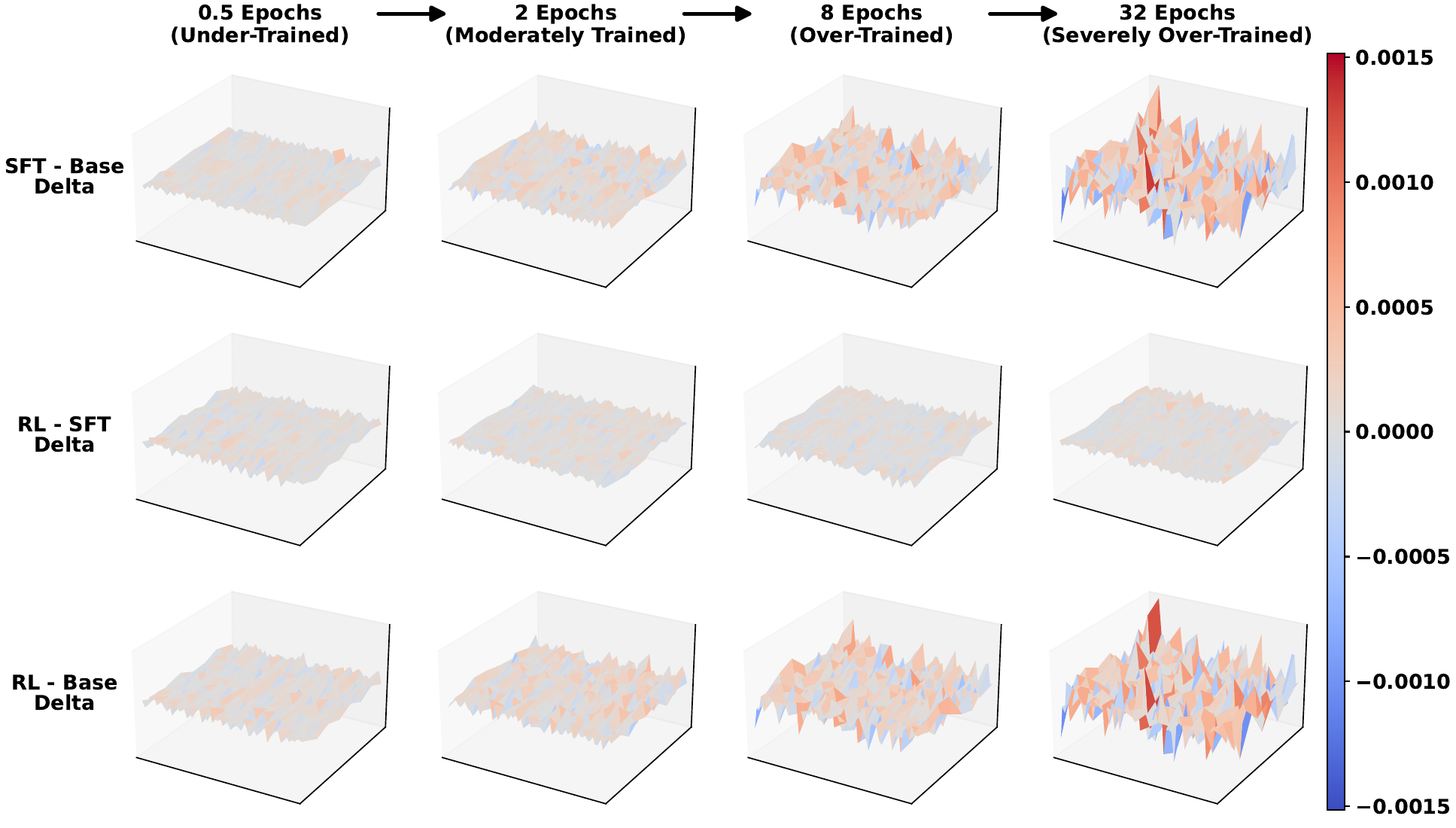}
\caption{Parameter changes of \texttt{layers.27.self\_attn.k\_proj.weight} induced by SFT and RL.}
\label{fig:param_change_l27_kproj}
\end{figure*}

To complement the per-module visualization in Section~\ref{sec:analysis_sft}, we provide additional layer-wise visualizations of the parameter shift induced by SFT and the subsequent RL stage. Figure~\ref{fig:param_change_l0_kproj} and Figures~\ref{fig:param_change_l27_vproj}-\ref{fig:param_change_l27_kproj} cover representative attention projections across early and late layers. The patterns are consistent with those reported in the main text: \over introduces sparse but extreme spikes that are concentrated in a small subset of parameters, while \mod produces relatively smooth and small-magnitude updates. Subsequent RL on \over is unable to undo these spikes, confirming that excessive SFT places the model in a region from which standard policy optimization cannot escape.

\end{document}